%% file: main.tex
\documentclass[10pt,twocolumn,letterpaper]{article}

\usepackage{cvpr}              


\definecolor{cvprblue}{rgb}{0.21,0.49,0.74}
\usepackage[pagebackref,breaklinks,colorlinks,allcolors=cvprblue]{hyperref}
\usepackage{multirow}
\usepackage{graphicx}
\usepackage{booktabs}

\usepackage{xcolor}
\usepackage{array}

\newcommand{\xhdr}[1]{{\noindent\bfseries #1}.}

\def\confName{CVPR}
\def\confYear{2025}

\newcommand{\ourmethod}{StdGEN}

\title{StdGEN: Semantic-Decomposed 3D Character Generation from Single Images}

\author{
    Yuze He$^{1,2}$\footnotemark[2]\quad
    Yanning Zhou$^1$\footnotemark[1]\quad
    Wang Zhao$^2$\quad
    Zhongkai Wu$^{1}$\quad
    Kaiwen Xiao$^1$\quad \\
    Wei Yang$^1$\quad
    Yong-Jin Liu$^2$\footnotemark[1]\quad
    Xiao Han$^1$ \\
    $^1$ Tencent AI Lab\quad
    $^2$ Tsinghua University
}

\begin{document}

\twocolumn[{%
\renewcommand\twocolumn[1][]{#1}%
\maketitle
\begin{center}
    \centering
    \vspace{-15pt}
    \captionsetup{type=figure}
    \includegraphics[width=1.0\linewidth]{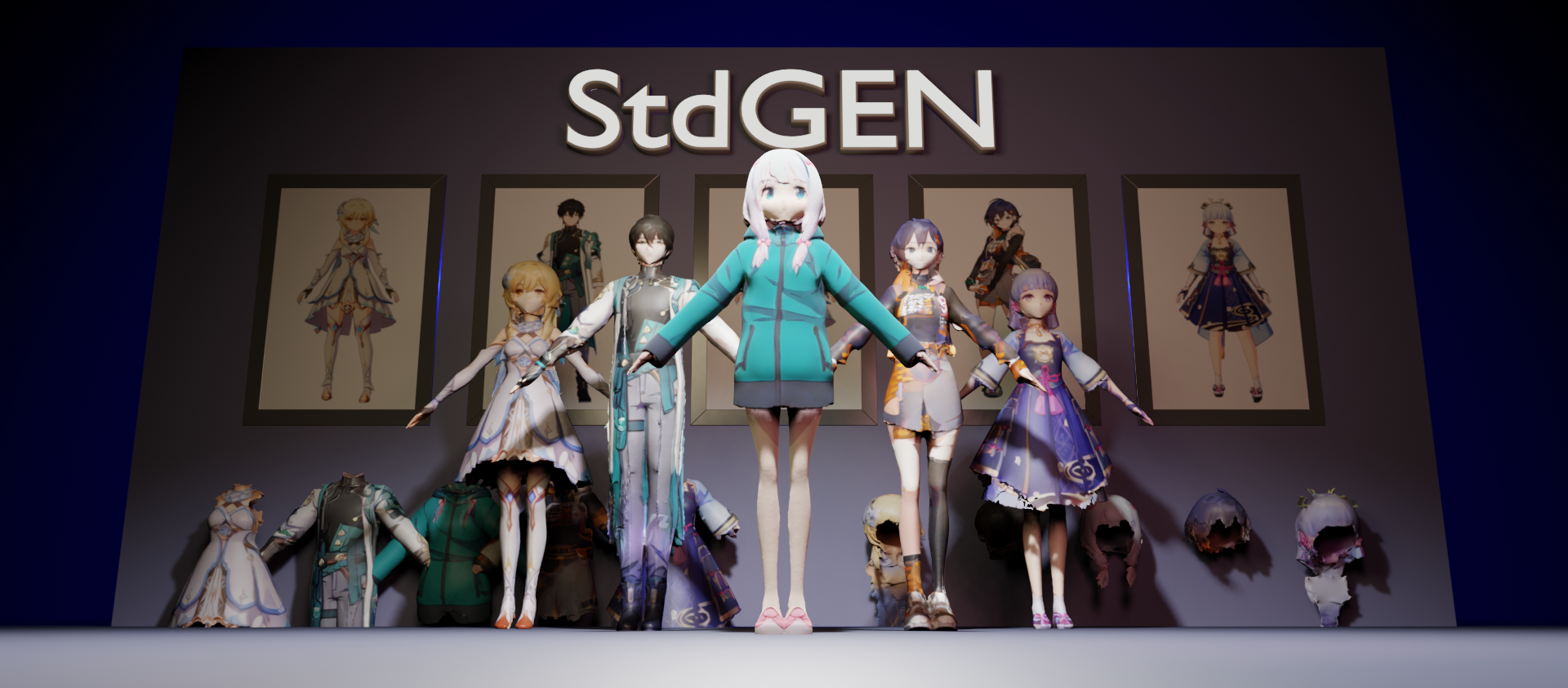}
    \captionof{figure}{Our \ourmethod\ generates high-quality, decomposed 3D characters from a single reference image.}
\end{center}%
}]

\renewcommand{\thefootnote}{\fnsymbol{footnote}}
\footnotetext[1]{Corresponding Authors: {\tt amandayzhou@tencent.com}, {\tt liuyongjin@tsinghua.edu.cn}}
\footnotetext[2]{Work done during an internship at Tencent AI Lab.}

\input{sec/0_abstract}
\vspace{-10pt}
\input{sec/1_intro}

\input{sec/2_related}
\input{sec/3_method}
\input{sec/4_experiments}
\input{sec/5_conclusion}
{
    \small
    \bibliographystyle{ieeenat_fullname}
    \bibliography{main}
}
\input{sec/X_suppl}

\end{document}

%% file: sec/0_abstract.tex
\begin{abstract}
We present \ourmethod, an innovative pipeline 
for generating semantically decomposed high-quality 3D characters from single images, enabling broad applications in virtual reality, gaming, and filmmaking, etc. Unlike previous methods which struggle with limited decomposability, unsatisfactory quality, and long optimization times, 
\ourmethod\ features decomposability, effectiveness and efficiency; i.e., it generates intricately detailed 3D characters with separated semantic components such as the body, clothes, and hair, in three minutes. At the core of \ourmethod\ is our proposed Semantic-aware Large Reconstruction Model (S-LRM), a transformer-based generalizable model that jointly reconstructs geometry, color and semantics from multi-view images in a feed-forward manner. A differentiable multi-layer semantic surface extraction scheme is introduced to acquire meshes from hybrid implicit fields reconstructed by our S-LRM. Additionally, a specialized efficient multi-view diffusion model and an iterative multi-layer surface refinement module are integrated into the pipeline to facilitate high-quality, decomposable 3D character generation. Extensive experiments demonstrate our state-of-the-art performance in 3D anime character generation, surpassing existing baselines by a significant margin in geometry, texture and decomposability. \ourmethod\ offers ready-to-use semantic-decomposed 3D characters and enables flexible customization for a wide range of applications.
Project page: \url{https://stdgen.github.io}
\end{abstract}

%% file: sec/1_intro.tex
\section{Introduction}
\label{sec:intro}
Generating high-quality 3D characters from single images has widespread applications in virtual reality, video games, filmmaking, etc. Beyond automatically creating a complete 3D character, there is an increasing demand for the ability to produce decomposable characters, where distinct semantic components like the body, clothes, and hair are disentangled. This decomposition allows for much easier editing, control, and animation of characters, greatly enhancing their usability across various downstream applications. 

However, creating such decomposable characters from single images is challenging, as each component may face issues such as occlusion, ambiguity, and inconsistencies in their interactions with other components. Existing methods for decomposable avatar generation primarily focus on realistic clothed human models, exploring disentangled 3D parametric~\cite{wang2023disentangled}, explicit~\cite{peng2024pica, pan2024humansplat}, or implicit~\cite{hong2022eva3d, huang2023avatarfusion, wang2024humancoser, dong2024tela} representations alongside various optimization techniques. These optimization approaches often employ score distillation loss~\cite{poole2022dreamfusiontextto3dusing2d} to leverage 2D generative priors, which leads to prolonged optimization times and the generation of coarse, high-contrast textures. Additionally, the dependence on parametric human models, such as SMPL-X~\cite{loper2023smpl}, is inadequate for virtual characters, which often exhibit exaggerated body proportions and complex clothing designs. 

To address these limitations, CharacterGen~\cite{peng2024charactergen} was developed to efficiently generate characters from single images using a multi-view diffusion model and large reconstruction model~\cite{hong2023lrm}. Despite showing impressive generation capabilities in various posed images, 
CharacterGen can only produce holistic avatars in watertight meshes with no decomposability. These meshes require significant manual labor to separate, edit, or animate, limiting their applicability. Moreover, the quality of the generated meshes is often unsatisfactory, particularly in finer details such as the character’s face and clothing, as shown in Fig.~\ref{fig:qual}. Therefore, efficiently generating high-quality, decomposable 3D characters remains an open challenge. 

In this work, we propose \ourmethod, an efficient and effective pipeline to generate decomposable, high-quality, A-pose 3D characters from single images of any pose. At its core is a novel Semantic-aware Large Reconstruction Model (S-LRM), which innovatively introduces semantic attributes to the original Large Reconstruction Model~\cite{hong2023lrm}, enabling efficient reconstruction of unified geometry, color and semantic fields. Moreover, a differentiable multi-layer semantic surface extraction scheme is proposed to extract decomposed semantic 3D surfaces from reconstructed implicit fields, empowering joint end-to-end training with explicit meshes. Together, S-LRM can efficiently generate semantic-decomposed, complete and consistent 3D surfaces from multi-view input images in a feed-forward manner.

Built upon S-LRM, \ourmethod\ comprises three key stages: multi-view diffusion, feed-forward reconstruction, and mesh refinement, each designed with technical innovations to enable effective decomposition. Specifically, given a single reference image with an arbitrary pose, a specialized efficient multi-view diffusion model first generates multiple A-pose RGB images and normal maps at different camera views. 
Next, the generated images are processed by our novel S-LRM, to reconstruct geometry, color, and semantics in a feed-forward manner. Once the coarse decomposed surface generation is obtained, we further refine the mesh quality through a proposed iterative multi-layer mesh refinement method, using diffusion-generated 2D images and normal maps as guidance.
As a result, \ourmethod\ can generate high-quality decomposable 3D characters from a single reference image within minutes. 

To facilitate the semantic decomposed training and testing, we further developed Anime3D++ dataset based on~\cite{peng2024charactergen}
, focusing on anime characters due to the wide internet presence, with finely annotated multi-view multi-pose semantic parts.
Our experiments demonstrate that \ourmethod\ surpasses all existing baselines, achieving state-of-the-art performance in both arbitrary and A-pose 3D character generation. Moreover, \ourmethod’s decomposable generation capability enables flexible customization, which can benefit numerous downstream applications. 
In summary, this work makes the following contributions:
\begin{itemize}
    \item We introduce a novel Semantic-aware Large Reconstruction Model (S-LRM), which jointly models geometry, color, and semantic information with efficient tri-plane feed-forward inference. An effective differentiable surface extraction scheme is proposed to facilitate the explicit multi-layer semantic surface reconstruction. 
    \item Building upon (1) our S-LRM, (2) an efficient multi-view diffusion model and (3) iterative multi-layer mesh refinement, we present \ourmethod, an efficient pipeline for high-quality decomposed 3D character generation, which for the first time enables semantic-decomposed 3D character creation from arbitrary-posed single images in minutes.
    \item Our method demonstrates superior performance over existing baselines, with its decomposed modeling unlocking potential for various downstream applications.
\end{itemize}

%% file: sec/2_related.tex
\section{Related Works}
\label{sec:related}

\begin{figure*}[htbp]
\centering
\vspace{-20pt}
\includegraphics[width=0.85\linewidth]{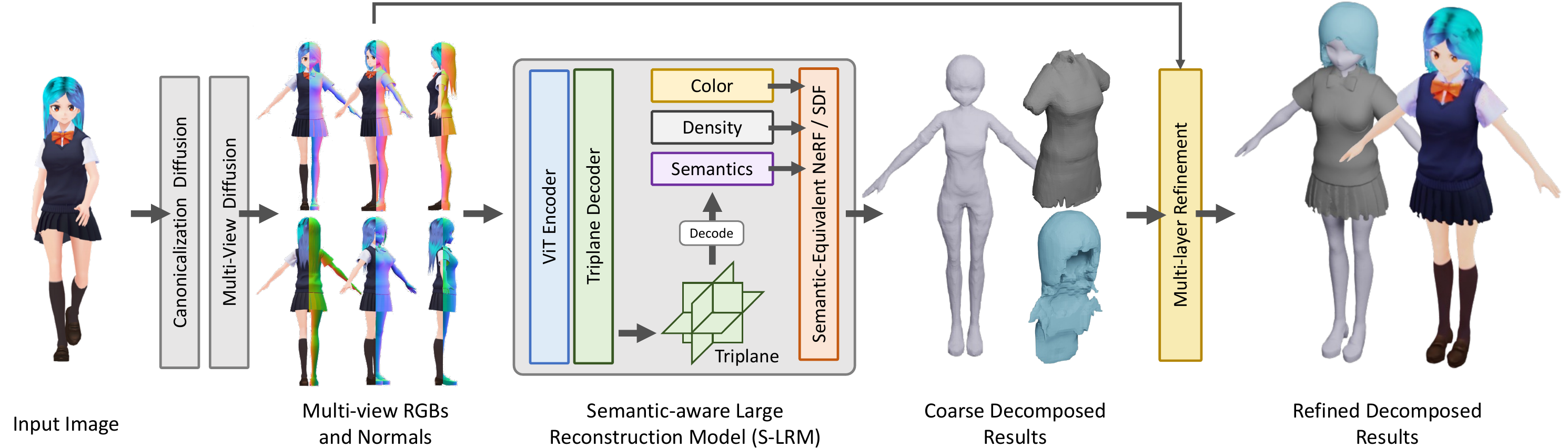}
\caption{The overview of our \ourmethod\ pipeline. Starting from a single reference image, our method utilizes diffusion models to generate multi-view RGB and normal maps, followed by S-LRM to obtain the color/density and semantic field for 3D reconstruction. Semantic decomposition and part-wise refinement are then applied to produce the final result.}
\label{fig:pipeline}
\vspace{-10pt}
\end{figure*}

\subsection{3D Generation}
To circumvent the need for extensive 3D assets during training, several approaches suggest lifting powerful 2D pre-trained diffusion models~\cite{dhariwal2021diffusion,nichol2021glide,rombach2022high,saharia2022photorealistic} for 3D generation.
The earliest works~\cite{poole2022dreamfusiontextto3dusing2d, wang2022scorejacobianchaininglifting} incorporate a pre-trained 2D diffusion model for probability density distillation using Score Distillation Sampling (SDS).
These approaches gradually optimize a randomly initialized radiance field~\cite{sun2022direct,chen2022tensorf,barron2022mip} with volume rendering, making it time-consuming to generate an object.
Later research continues to enhance the aesthetics and accuracy of 3D content generation~\cite{lin2023magic3d, chen2023fantasia3d, tsalicoglou2023textmesh, shen2021dmtet, wang2023prolificdreamer} and further investigate different application scenarios~\cite{haque2023instruct,shao2023control4d,singer2023text,raj2023dreambooth3d}.
However, relying solely on 2D priors for 3D generation often leads to poor geometry representation, e.g., multi-faced Janus problem, due to the challenges in controlling precise viewpoints through text prompts.
The large-scale 3D datasets, e.g. Objaverse~\cite{deitke2023objaversexluniverse10m3d}, unlock the possibility of imposing 3D priors to the model. 
Several works utilize view-consistent images to fine-tune the diffusion model.  
Zero-1-to-3~\cite{liu2023zero} integrates 3D priors into 2D stable diffusion by fine-tuning the pre-trained model for novel view synthesis (NVS). To further enhance the multi-view consistency, several recent works~\cite{shimvdream, long2024wonder3d, liu2023syncdreamer, huang2024epidiff} propose synchronously generating multi-view images in a single generation process and achieving constraints in 3D place through feature interaction in attention mechanism.
Besides, the 3D native generation method shows powerful geometric generation ability~\cite{zhang2024clay,li2024craftsman,lu2024direct2}. 
However, the ability of these methods to follow instructions is typically moderate; therefore, they face challenges in achieving the desired outcomes in scenarios requiring precise restoration of reference images, e.g., 3D character generation.

\subsection{Large Reconstruction Model}
Large Reconstruction Model~(LRM)~\cite{hong2023lrm} leverages the transformer-based model to map the single image feature to implicit tri-plane representation. 
Instant3D~\cite{liinstant3d} extends LRM by feeding multi-view images instead of a single image.
LGM~\cite{tang2024lgm}, GRM~\cite{xu2024grm} and GS-LRM~\cite{gslrm2024} replace the 3D representation to 3D Gaussians, embracing its efficiency in rendering and low memory consumption.
InstantMesh~\cite{xu2024instantmesh} and CRM~\cite{wang2024crm} explicitly model the geometry by equipping the generative pipeline with FlexiCubes~\cite{shen2023flexible}, achieving high-quality surface extraction and high rendering speed. 
The following works further explored applying advanced model architecture~\cite{zhang2024geolrmgeometryawarelargereconstruction} or 3D presentation~\cite{chen2024laraefficientlargebaselineradiance, cui2024lam3dlargeimagepointcloudalignment}, aiming to improve the efficiency, realism, and generalization of reconstruction.
Integrating with multi-view diffusion models, these LRMs can achieve text-to-3D generation or single image-to-3D generation.
Yet all these methods typically produce holistic models.
In contrast, our method generates semantically decomposed characters, making downstream processing such as editing and animation much more efficient.

\subsection{3D Character Generation}

3D character generation is a challenging problem due to its high precision requirements and the scarcity of data.
One line of work leverages 3D-aware GANs to model the distribution of digital humans
~\cite{bergman2023generativeneuralarticulatedradiance, hong2022eva3d, jiang2022humangengeneratinghumanradiance, zhang2022avatargen3dgenerativemodel, noguchi2022unsupervisedlearningefficientgeometryaware}.
Recently the SDS-based methods have shown the possibility of generating a variety of stylized characters~\cite{cao2023dreamavatartextandshapeguided3d, huang2023dreamwaltzmakescenecomplex, wang2023disentangled, dong2024tela, kim2024gala}, yet it suffers from the long optimization times and the difficulty of meticulous style control.
Frankenstein~\cite{yan2024frankenstein} concentrates on producing decomposed, textureless 3D meshes based on 2D layouts, restricting the potential for achieving high-fidelity reconstruction from the reference image.
CharacterGen~\cite{peng2024charactergen} calibrates input poses to canonical multi-view images via an image-conditioned multi-view diffusion model, followed by LRM for 3D character reconstruction and multi-view texture back projection, but still exhibits limited geometry and texture quality.
Our approach, in contrast, employs a semantic-aware, feed-forward paradigm that generates high-quality, decomposable characters using only one forward pass from an arbitrary reference image, providing significant efficiency and quality improvement.

%% file: sec/3_method.tex
\section{Method}
\label{sec:method}

The proposed \ourmethod\ begins with multi-view canonical character generation (Sec.~\ref{subsec:multiview}) from a reference image.
To reconstruct a decomposable 3D character from multi-view images, we extend the LRM with a semantic field, enabling semantic-based layered generation (Sec.~\ref{subsec:lrm}).
Finally, a multi-level refinement process is designed to enhance the results, improving the geometric structure and providing more detailed textures (Sec.~\ref{subsec:refine}). An overview of the \ourmethod\ pipeline is shown in Fig.~\ref{fig:pipeline}.

\subsection{Multi-view Generation and Canonicalization}
\label{subsec:multiview}
Given a reference character image, our objective is to generate a set of multi-view images and corresponding normal maps that maintain 3D consistency. This process involves two steps: Canonicalizing an arbitrary reference image into an A-pose character and generating multi-view RGBs and normals from the A-pose representation.

\xhdr{Arbitrary Reference to A-pose Conversion}
Directly reconstructing a 3D character model in arbitrary poses can be affected by self-occlusion from different viewpoints. Therefore, we design to first canonicalize the character in 2D space.
We select the A-pose as the target representation, as it is more conducive to subsequent multi-view generation, reconstruction, and applications like rigging. 
Following previous works~\cite{hu2024animate, peng2024charactergen}, we employ the Stable Diffusion~\cite{rombach2022high} model, augmented with a ReferenceNet, to translate a 2D arbitrary posed reference image to A-pose image. The ReferenceNet helps to preserve the reference identity in the resulting image. We refer to~\cite{peng2024charactergen} for more details.

\xhdr{Multi-view RGBs and Normals Generation}
This stage provides comprehensive information for the subsequent S-LRM, with multi-view consistent normals crucial for capturing rich surface details in mesh refinement
rather than relying on independent normal predictions.
We adapt Era3D~\cite{li2024era3d} to generate high-resolution RGBs and normals simultaneously. Using memory-efficient row-wise attention across views and between RGB and normal maps, we ensure fine-grained spatial consistency and enable higher output resolutions.
Specifically, it takes a single A-pose image as input, producing orthographic projections of six viewpoints (elevation $0^\circ$, azimuth ${-90^\circ, -45^\circ, 0^\circ, 45^\circ, 90^\circ, 180^\circ}$).
For sharper character details, we increased the resolution through a progressive training approach, ultimately reaching 1024.

Compared with CharacterGen~\cite{peng2024charactergen}, our choice can simultaneously generate high-resolution, multi-view consistent normal maps for mesh refinement. 
Besides, the two-step design allows for improved editing in the 2D A-pose space, facilitating the generation of decomposed characters for enhanced 3D editing applications.

\subsection{Semantic-aware Large Reconstruction Model}
\label{subsec:lrm}
Once obtaining multi-view images,~\cite{xu2024instantmesh, peng2024charactergen} use transformer based sparse-view Large Reconstruction Model~(LRM) to reconstruct a holistic 3D mesh.
In contrast, we aim to generate characters with decomposed components, including the minimal-clothed human model, external clothing, and hair, to produce 3D character models that are not only visually accurate but also functionally versatile for various uses in 3D games and animation pipelines.
To achieve this goal, we proposed the Semantic-aware Large Reconstruction Model (S-LRM), which extends NeRF/SDF to simultaneously encode semantics, appearance and geometry information in a feed-forward manner.

As shown in Fig.~\ref{fig:pipeline}, the S-LRM takes multi-view images into tokens and then feeds into a transformer-based image-to-triplane decoder to output a triplane representation.
The triplane features are decoded to obtain 
semantic, color, density/SDF information.
To enhance the training efficiency and reconstruction quality, following InstantMesh~\cite{xu2024instantmesh}, we first utilize triplane NeRF representation to render 2D images with volume rendering for training. We then switch to FlexiCubes~\cite{shen2023flexible} to extract explicit mesh from triplane-decoded SDF grid, and use direct mesh rasterization to render images. However, to obtain semantic-decomposed surface reconstruction, both NeRF and SDF implicit representations must be capable of rendering distinct semantic layers into images or extracting separate semantic surfaces using FlexiCubes in a differentiable manner.
To achieve that, a novel semantic-equivalent NeRF/ SDF is proposed to extract character parts by specific semantics.

\begin{figure}[tb]
\centering
\includegraphics[width=0.95\linewidth]{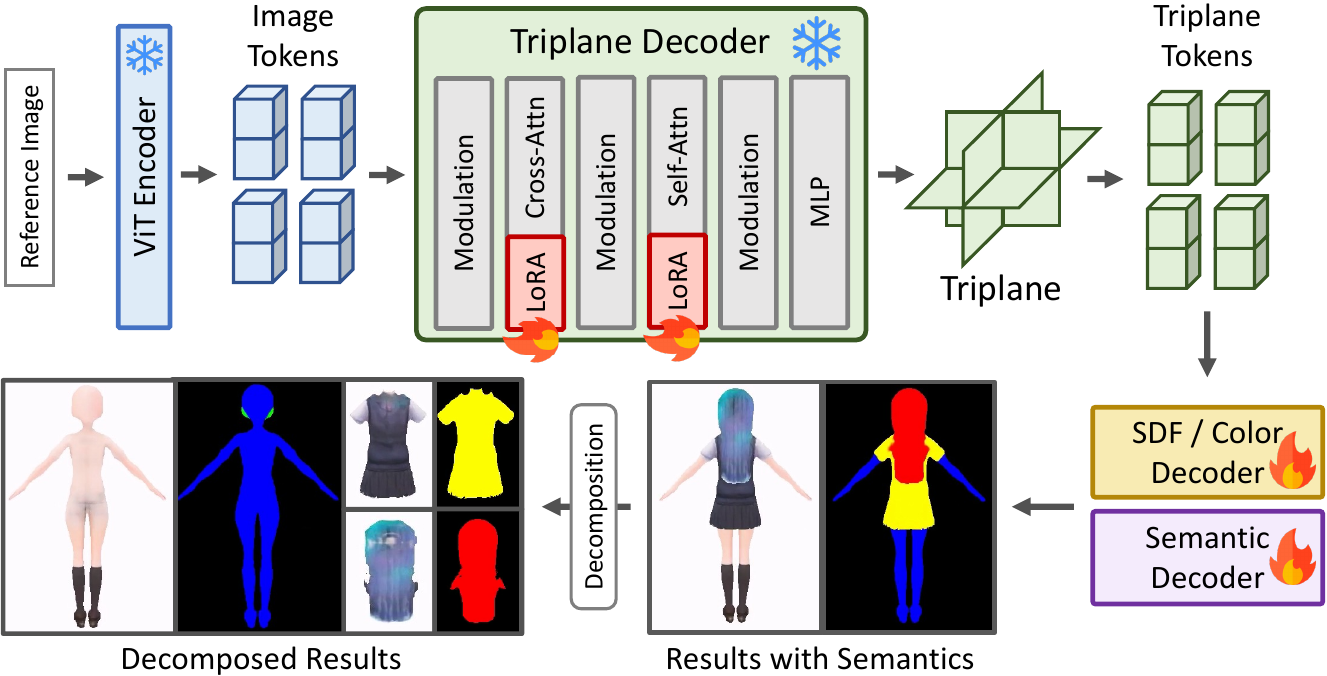}
\caption{Demonstration of the structure and intermediate outputs of our semantic-aware large reconstruction model (S-LRM). }
\label{fig:lrm}
\vspace{-15pt}
\end{figure}

\xhdr{Semantic-equivalent NeRF and SDF}
NeRF represents a 3D scene by spatial-variant volume densities with colors\footnote{We ignore the view-dependent effects to simplify the discussion.}.
We extend it with a semantic field, and model them as a learnable function $F_\Theta$ that takes sampled point location $\mathbf{x} = (x; y; z)$ as inputs, and outputs color $c$, density $\sigma$ and semantic distribution $s$ as:
$(\sigma, c, s) = F_\Theta(\mathbf{x})$.

To render per-pixel color $\hat{C}(\mathbf{r})$, a series of 3D points are sampled along the ray $\mathbf{r}$, and the pixel color is computed by integrating the sampled densities $\sigma_i$ and colors $c_i$ using the volume rendering equation with:
\begin{align} 
\hat{C}(\boldsymbol{r}) &=\sum_{i=1}^N T_{i}\alpha_i c_i,\ T_{i}=\prod_{j=1}^{i-1}(1-\alpha_j)
\end{align}
where $\alpha_i=(1-\exp (-\sigma_i \delta_i))$, $\delta_i=t_{i+1}-t_i$ is the alpha value of samples and distance between adjacent samples. 

Given the probability $p_{s,i}$ of semantic $s$ at location $i$, the pixel color $\hat{C}_s(\mathbf{r})$ under semantic $s$ can be calculated as:
\begin{align} 
\hat{C}_s(\boldsymbol{r}) &=\sum_{i=1}^N T_{s,i}p_{s,i}\alpha_i c_i,\ T_{s,i}=\prod_{j=1}^{i-1}(1-\alpha_jp_{s,j})
\end{align} 

If the probability of a certain semantic at a given location is zero, it should be considered fully transparent under the current semantic category. Furthermore, given that a position is known to be opaque, the probability of the current semantic should be linear to the final equivalent transparency.

\begin{figure}[tb]
\includegraphics[width=1.0\linewidth]{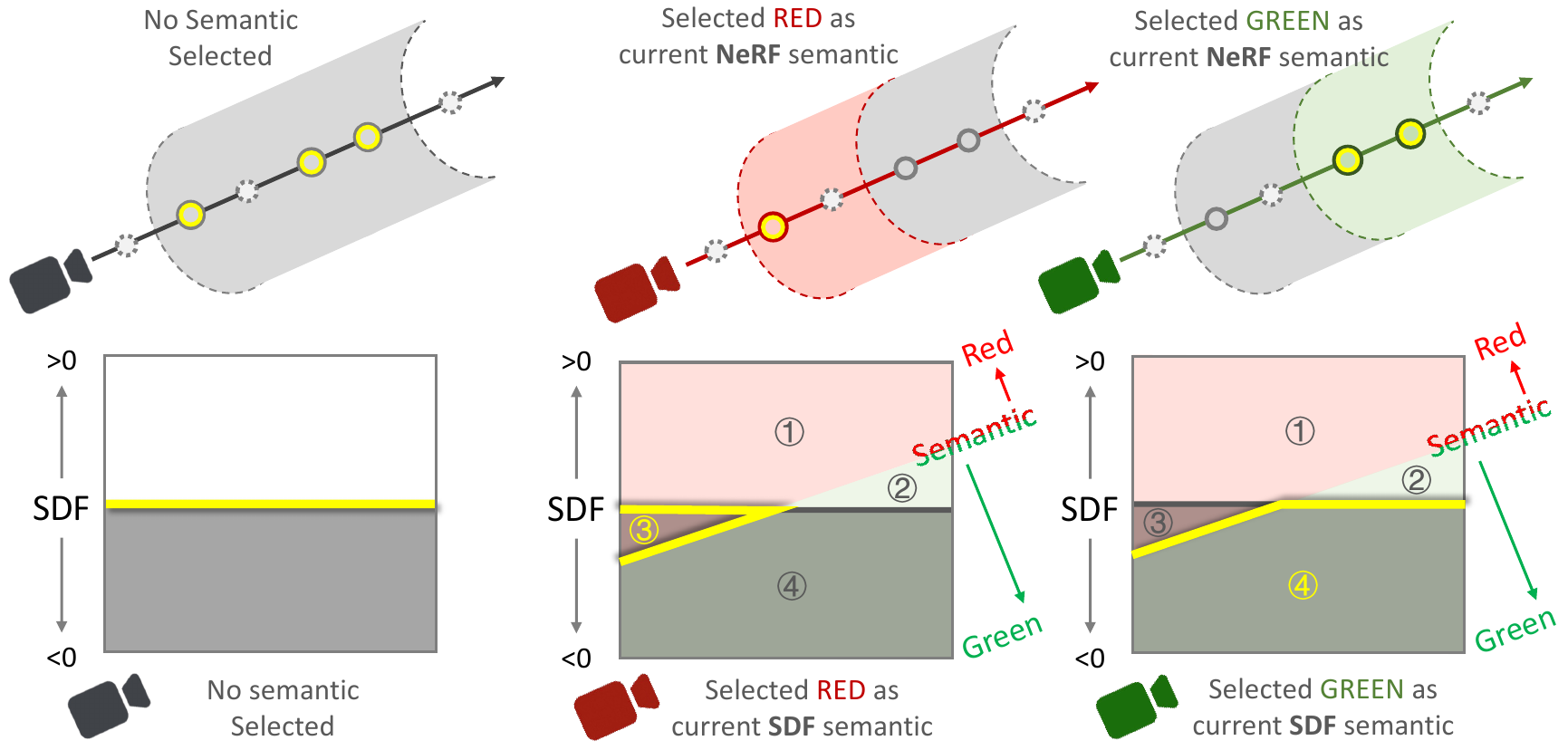}
\caption{Our semantic-equivalent NeRF and SDF extraction scheme (shown in yellow color).}
\label{fig:semantic}
\vspace{-10pt}
\end{figure}

Unlike NeRF, SDF does not incorporate the concept of transparency.
Instead, positive/negative values represent points outside/inside the surface. 
Consequently, semantic probabilities cannot be directly applied to SDF for the mesh part extraction. 
Upon analysis, the extraction of a semantic-equivalent SDF should adhere to the following principles:

\begin{enumerate}
    \item The zero value of the original SDF serves as a hard constraint. When the original SDF is positive, the equivalent SDF should also be positive;
    \item When the original SDF is negative, the equivalent SDF should be zero at the boundaries where the maximum of relevant semantics transits;
    \item At locations where the original SDF equals zero, but the probability of the current semantic is not the highest among all semantics, the equivalent SDF should not only maintain its sign but also be greater than zero.
\end{enumerate}

Based on these principles, we propose the following formula for constructing the equivalent SDF:
\vspace{-15pt}

\begin{align}
    f_{i,s}=\max(f_i, (\max_{r\neq s}p_{i,r})-p_{i,s} ),
\end{align}
Where $f_{i}$, $f_{i,s}$ are original SDF and equivalent SDF of semantic $s$ at location $i$, respectively.
Fig.~\ref{fig:semantic} illustrates our method's scheme. For red semantics, only region 3 is selected, as regions 1, 2 (SDF$>$0) and region 4 (non-red) are discarded. Similarly, when the green is chosen, region 4 is correctly extracted.
This formulation ensures correct decomposition by specific semantics and is fully compatible with subsequent FlexiCubes mesh extraction. In this way, we can differentiably extract multi-layer semantic surfaces from S-LRM's outputs, greatly facilitating the LRM training and downstream optimization.

\xhdr{Model Structure}
Our S-LRM's core structure is derived from InstantMesh~\cite{xu2024instantmesh} with a ViT encoder, an image-to-triplane transformer, and the feature decoder.
Several enhancements are made to this foundation to better suit our objectives, with the semantic decoder being a key addition. 
This component mirrors the structure of the density/color decoder and is designed to generate the semantic field in 3D space. 
For memory efficient training, we incorporate LoRA~\cite{hu2021lora, qi2024tailor3d} structures into all linear layers inside both self-attention and cross-attention blocks.

\xhdr{Multiple semantic level supervision}
Current LRMs typically rely solely on 2D supervision, which limits their ability to generate information about objects' internal structures under occlusion; 3D supervision would be effective but often too resource-intensive. 
To address this, we propose an effective supervision that jointly learns semantics and colors, enabling the acquisition of a 3D semantic field and internal character information using only 2D supervision. 

\noindent\textit{Stage 1: Training on NeRF with Single-layer Semantics. }
In this initial stage, we train on the triplane NeRF representation.
We initialize the model with the pre-trained InstantNeRF, training the newly added LoRA in all attention blocks' linear layers and the newly introduced semantic decoder.
We train it under the image, mask and semantic loss:

\vspace{-15pt}
\begin{align}
    & \hat{\mathcal{S}}(\boldsymbol{r}) =\sum_{i=1}^N T_ip_i\alpha_i,\
    \mathcal{L}_{\text{sem}} = \sum_k CE(\hat{\mathcal{S}_k}, \mathcal{S}^{gt}_k)
    \\ & \mathcal{L}_1 = \mathcal{L}_{\text{mse}} + \lambda_{\text{lpips}} \mathcal{L}_{\text{lpips}} + \lambda_{\text{mask}}\mathcal{L}_{\text{mask}} + \lambda_{\text{sem}}\mathcal{L}_{\text{sem}}
\end{align}
$\hat{\mathcal{S}}$ is the semantic map calculated by the probabilities $p_{i}$ from semantic decoder's output through a softmax layer.
$\hat{\mathcal{S}_k}, \mathcal{S}^{gt}_k$ denotes the $k$-th view of rendered and ground-truth semantic maps, and $CE$ denotes the cross-entropy function.

\newcolumntype{C}[1]{>{\centering\arraybackslash}m{#1}}

\begin{table*}[htb]
    \centering
    \resizebox{0.78\linewidth}{!}{
    \begin{tabular}{clcccc|cccc}
        \toprule
        & & \multicolumn{4}{c}{A-pose Conditioned Input} & \multicolumn{4}{c}{Arbitrary-pose Conditioned Input} \\
        \cmidrule(lr){3-6} \cmidrule(lr){7-10}
        & & SSIM$\uparrow$ & LPIPS$\downarrow$ & FID$\downarrow$ & CLIP Similarity$\uparrow$ & SSIM$\uparrow$ & LPIPS$\downarrow$ & FID$\downarrow$ & CLIP Similarity$\uparrow$ \\
        \midrule
        \multirow{5}{*}{\textbf{\shortstack{Multi-view \\ Comparisons \\ in 2D}}} & SyncDreamer~\cite{liu2023syncdreamer}   & 0.870	& 0.183	& 0.223	& 0.864	& 0.845	& 0.217	& 0.328 &	0.839 \\
        & Zero-1-to-3~\cite{liu2023zero}       & 0.865	& 0.172	& 0.500	& 0.885	& 0.842	& 0.209	& 0.481 & 	0.878 \\
        & Era3D~\cite{li2024era3d}         & 0.876	& 0.144	& 0.095	& 0.908	& 0.842	& 0.195	& 0.094 & 	0.900 \\
        & CharacterGen~\cite{peng2024charactergen}  & 0.886	& 0.119	& 0.063	& 0.928	& 0.871	& 0.139	& 0.056 & 	0.919 \\
        & Ours          & \textbf{0.958}	& \textbf{0.038}	& \textbf{0.004}	& \textbf{0.941}	& \textbf{0.920}	& \textbf{0.071}	& \textbf{0.014} &	\textbf{0.935} \\
        \midrule
        \multirow{8}{*}{\textbf{\shortstack{Character \\ Comparisons \\ in 3D}}} & Magic123~\cite{qian2023magic123}      & 0.886	& 0.142	& 0.192	& 0.887	& 0.849	& 0.197	& 0.256 &	0.862 \\
        & ImageDream~\cite{wang2023imagedream}    & 0.856	& 0.171	& 0.846	& 0.836	& 0.823	& 0.218	& 0.875 &	0.818 \\
        & OpenLRM~\cite{openlrm}       & 0.889	& 0.151	& 0.406	& 0.878	& 0.863	& 0.191	& 0.707 &	0.844 \\
        & LGM~\cite{tang2024lgm}           & 0.876	& 0.151	& 0.282	& 0.902	& 0.838	& 0.203	& 0.480 &	0.884 \\
        & InstantMesh~\cite{xu2024instantmesh}   & 0.888	& 0.126	& 0.107	& 0.906	& 0.846	& 0.202	& 0.285 &	0.886 \\
        & Unique3D~\cite{wu2024unique3d}      & 0.889	& 0.136	& 0.030	& 0.919	& 0.856	& 0.190	& 0.042 &	0.903 \\
        & CharacterGen~\cite{peng2024charactergen}  & 0.880	& 0.124	& 0.081	& 0.905	& 0.869	& 0.134	& 0.119 &	0.901 \\
        & Ours          & \textbf{0.937}	& \textbf{0.066}	& \textbf{0.010}	& \textbf{0.941}	& \textbf{0.916}	& \textbf{0.084}	& \textbf{0.011} &	\textbf{0.936} \\
        \bottomrule
    \end{tabular}
    }
    \vspace{-5pt}
    \caption{Quantitative comparison of A-pose and arbitrary pose inputs for 2D multi-view generation and 3D character generation.}
    \label{tb:main}
    \vspace{-12pt}
\end{table*}

\noindent\textit{Stage 2: Training on NeRF with Multi-layer Semantics. }
Having learned robust surface semantic information in the first stage, we aim to learn the 3D character's internal semantic and color information.
We hierarchically supervise from outside to inside according to the spatial relationship of different semantic parts, by masking specific semantics during rendering and supervising with corresponding 2D ground truth.
Assuming we aim to preserve a set of semantics $\{P_s\}$, we can render the image and semantic map under current conditions as follows:
\vspace{-15pt}

\begin{align}
    \hat{C}_P(\boldsymbol{r}) &=\sum_{i=1}^N T_{P,i}\alpha_i c_i\sum_{s\in P}p_{s,i}, \\
    \hat{\mathcal{S}}_P(\boldsymbol{r}) &=\sum_{i=1}^N T_{P,i}\alpha_i p_i\sum_{s\in P}p_{s,i}, \\
    \mbox{where }\ T_{P,i} &=\prod_{j=1}^{i-1}(1-\alpha_j\sum_{s\in P}p_{s,j}),
\end{align}
The loss function is defined as:
\begin{align}
\mathcal{L}_2 & = \mathcal{L}_{\text{mse},P} + \lambda_{\text{lpips}} \mathcal{L}_{\text{lpips},P} + \lambda_{\text{mask}}\mathcal{L}_{\text{mask},P}\nonumber \\
&+ \lambda_{\text{sem}}\sum_k CE(\hat{\mathcal{S}}_{P,k}, \mathcal{S}^{gt}_{P,k})
\end{align}
This decomposed training approach enables our S-LRM to simultaneously learn color and semantic information for the surface and the object's interior, thus achieving feed-forward 3D content decomposition and reconstruction.

\noindent\textit{Stage 3: Training on Mesh with Multi-layer Semantics. }
We switch to mesh representation~\cite{shen2023flexible}
for efficient high-resolution training.
We then extract the equivalent SDF via:
\begin{align}
    f_{i,P}=\max(f_i, (\max_{s\notin P}p_{i,s}-\max_{s\in P}p_{i,s}) ),
\end{align}
Subsequently, we input the equivalent SDF into FlexiCubes to obtain the mesh, render the image and semantic map, and supervise using the following loss function:
\vspace{-12pt}

\begin{align}
    \mathcal{L}_3 &= \mathcal{L}_2 + \lambda_{\text{normal}} \sum_k M^{gt}_P\otimes\left(1 - \hat{N}_{P,k}\cdot N_{P,k}^{gt}\right) \nonumber\\
&+ \lambda_{\text{depth}} \sum_{k} M^{gt}_P\otimes\left\|\hat{D}_{P,k}-D_{P,k}^{gt}\right\|_1 
 +\lambda_{\text{dev}}\mathcal{L}_{\text{dev}}
\end{align}
where $\hat{D}_{P,k}$, $\hat{N}_{P,k}$, denotes the rendered depth and normal; $D_{P,k}^{gt}$, $ N_{P,k}^{gt}$ and $M_P^{gt}$ denote the ground truth depth, normal, and mask of the $k$-th view under semantic set $P$, respectively; $\mathcal{L}_{\text{dev}}$ denotes the deviation loss of FlexiCubes.

\subsection{Multi-layer Refinement}
\label{subsec:refine}
Given the limitations in geometric and texture detail achievable by large reconstruction models, further optimization of the mesh post-reconstruction is necessary. Recent methods~\cite{wu2024unique3d, li2024craftsman} utilizing high-resolution normal maps for mesh optimization have shown promising results, albeit primarily for holistic meshes. We propose an iterative optimization mechanism for multi-layer mesh refinement.

To ensure thorough optimization at each level, we employ a staged approach following our S-LRM: Initially, we extract different parts of the mesh by specifying various semantics and optimize only the base minimal-clothed human model; upon completion, we overlay the clothing mesh, fixing the base and optimizing solely the clothing component; finally, we add the hair mesh, fixing the previous two layers and optimizing only the hair.
The optimization process is guided by the multi-view normal maps generated earlier through diffusion models, each optimization step involves a differentiable rendering of the mesh to compute losses and gradients, followed by vertex adjustments based on these gradients, and re-mesh operations including edge collapse, split and flips. The loss function is defined as follows:
\vspace{-12pt}

\begin{align}
    \mathcal{L}_{r1}=\ & \lambda_{\text{mask}}' \sum_k || \hat{M}_k-M_k^{\text{pred}} ||_2^2 +\ \lambda_{\text{col}} \mathcal{L}_{\text{col}} \nonumber \\
    +\ & \lambda_{\text{normal}}' M_k^{\text{pred}} \otimes \sum_k || \hat{N}_k-N_k^{\text{pred}} ||_2^2
\end{align}
where $\hat{M}_k$, $\hat{N}_k$ are rendered masks and normal maps, $M_k^{\text{pred}}$, $N_k^{\text{pred}}$ are diffusion-generated masks and normal maps under $k$-th view, respectively. $\mathcal{L}_{\text{col}}$ is the collision loss modified from \cite{peng2024pica} to ensure outer-layer mesh in the outer normal direction of inner-layer mesh:
\vspace{-15pt}

\begin{align}
    \mathcal{L}_{\text{col}}=\frac{1}{n}\sum_{i=1}^n\max{((v_j-v_i)\cdot n_j, 0)^3}
\end{align}
where $v_i$ represents the $i$-th vertex of the outer-layer mesh, $v_j$ is its nearest neighbor of $v_i$ on the inner-layer mesh, and $n_j$ denotes the normal vector associated with $v_j$.
Upon completing the optimization process, the mesh undergoes an additional ExplicitTarget Optimization phase, similar to that employed in Unique3D~\cite{wu2024unique3d}. This stage aims to eliminate multi-view inconsistencies and further refine the geometry. Finally, the optimized meshes are colorized by the back projection of the multi-view images.

%% file: sec/4_experiments.tex
\section{Experiments}
\label{sec:experiments}

\begin{figure*}[htbp]
\centering
\includegraphics[width=0.85\linewidth]{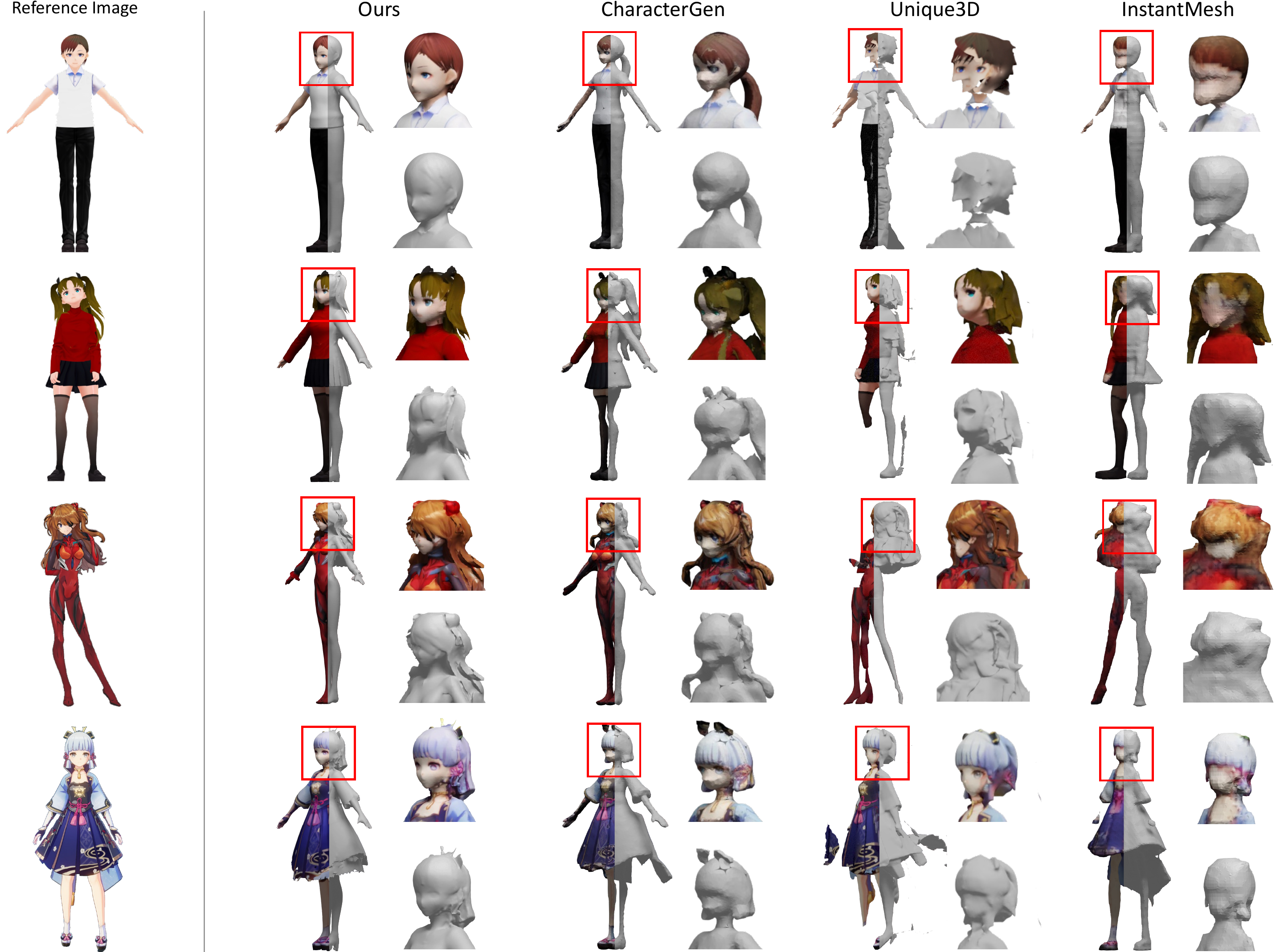}
\vspace{-5pt}
\caption{Qualitative comparisons on geometry and appearance of generated 3D characters.}
\label{fig:qual}
\vspace{-10pt}
\end{figure*}

\subsection{Anime3D++ Dataset}
We develop the Anime3D++ Dataset to meet the quality and decomposability requirements based on the original Anime3D~\cite{peng2024charactergen} Dataset. We initially gathered around 14,000 3D anime character models from VRoid-Hub, then filtered these down to 10,811 high-quality models, standardized in A-pose with arms angled 45° downward. Renderings include RGB images, depth, normal, and semantic maps for layered supervision. Each character is rendered in full, only a base minimal-clothed human model and a base human model plus clothing. Please refer to the supplementary materials (Sec.~E) for more details.

\subsection{Results and Comparisons}
Given that current methods cannot generate layered 3D models from a single arbitrary character image, we compare our non-layered generation results with other methods on the test split of our Anime3D++ dataset. We further decoupled the pose canonicalization component to ensure fairness, conducting both 2D and 3D comparisons for arbitrary pose and A-pose character reference inputs. When using A-pose inputs, all methods are compared against A-pose ground truth. For arbitrary pose inputs, we follow CharacterGen's settings, comparing our method and CharacterGen (both capable of canonicalization) against the A-pose ground truth, while other methods are compared under the original pose ground truth.
Generation quality and fidelity are evaluated using SSIM~\cite{wang2004image}, LPIPS~\cite{zhang2018perceptual}, and FID. We also compute CLIP~\cite{radford2021learning} cosine similarity between the front ground-truth image and the multi-views or 3D renderings.

\xhdr{Quantitative Results}
For 2D multi-view generation results, we compare our method with Zero-1-to-3~\cite{liu2023zero}, SyncDreamer~\cite{liu2023syncdreamer}, Era3D~\cite{li2024era3d}, and CharacterGen~\cite{peng2024charactergen}. For 3D Character Generation results, we compare with Magic123~\cite{qian2023magic123}, ImageDream~\cite{wang2023imagedream} (SDS-based optimization); OpenLRM~\cite{hong2023lrm,openlrm}, LGM~\cite{tang2024lgm}, InstantMesh~\cite{xu2024instantmesh} (feed-forward methods); Unique3D~\cite{wu2024unique3d} (direct mesh reconstruction); and CharacterGen. 3D Results are uniformly rendered as eight equidistant images at elevation=0 and aligned using horizontal mask registration.

As shown in Tab.~\ref{tb:main}, Our method performs better in both standard and arbitrary pose settings.
Existing 2D multi-view generation methods often struggle to preserve adequate geometric and appearance information in generated images. Among 3D methods, SDS-based approaches typically exhibit blurred geometry and suffer from the Janus Problem, and feed-forward methods generally lack geometric and texture precision.
Unique3D achieves high metrics due to high-resolution supervision but suffers from unstable mesh initialization, impacting visual quality.
CharacterGen demonstrates a notable advantage to other methods in arbitrary pose settings but the advantage diminishes in A-pose, showing effective pose canonicalization but limited reconstruction ability.
In contrast, our method consistently achieves superior results across all cases.

\begin{figure}[t]
\centering
\includegraphics[width=1\linewidth]{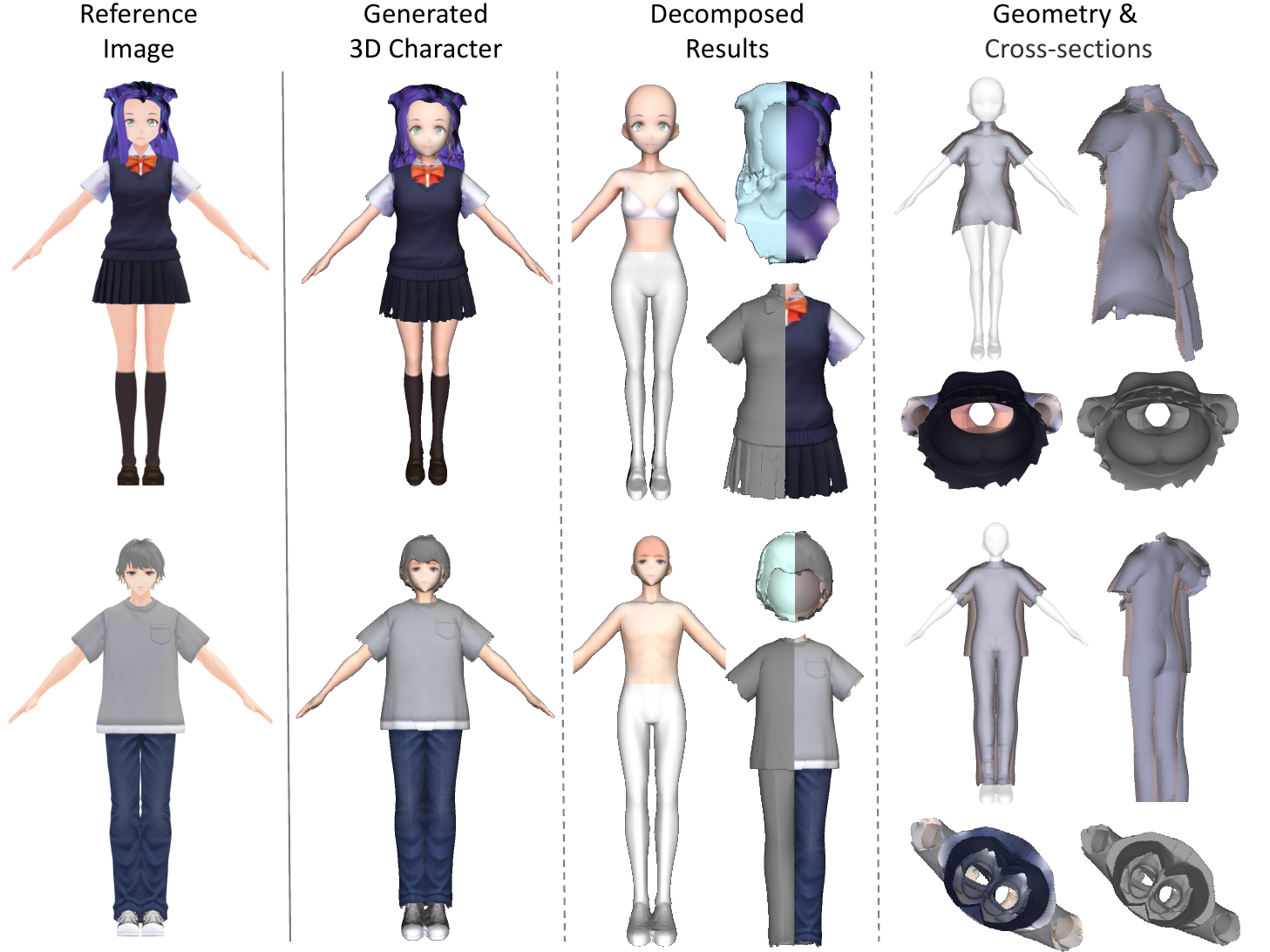}
\caption{Decomposed outputs of our method, presented in texture, mesh, and cross-section.}
\label{fig:layers}
\vspace{-10pt}
\end{figure}

\xhdr{Qualitative Results}
Fig.~\ref{fig:qual} reveals several limitations across different methods. InstantMesh's results were constrained by grid resolution, leading to insufficient texture precision. While Unique3D achieved higher resolution, its heavy reliance on depth-based mesh initialization made it susceptible to geometric collapse under inaccurate depth estimations. CharacterGen exhibited low geometric and texture fidelity despite employing multi-view back-projection, and frequently produced visually disruptive black artifacts. In contrast, our method demonstrated superior geometric accuracy and texture fidelity performance.

\xhdr{Decomposed Results}
Fig.~\ref{fig:layers} illustrates our method's capability to reconstruct decomposed characters from single reference images, organized from left to right as follows: the input reference image, the reconstructed 3D character (applied shading for better visualization), the semantically decomposed components and geometries. The reconstructed meshes demonstrate high geometric precision, while the semantic decomposition is also accurate, successfully decoupling the base human model, clothing, and hair. This level of decomposition represents a significant advancement in character reconstruction from single images.
The rightmost figure presents a cross-sectional view of the clothing, revealing that our reconstructed clothing is entirely internally hollow. This substantially enhances the potential for integration with downstream applications like realistic physics simulations and easier rigging.

\xhdr{User Study}
To comprehensively evaluate our method's performance, we conducted a user study involving 28 volunteers. The study utilized 16 randomly selected 3D meshes and their corresponding reference images, drawn from both web-collected images and the Anime3D++ test split. Participants were asked to compare the randomly shuffled results of four approaches based on overall quality, fidelity, geometric quality, and texture quality. For each comparison group, participants identified the best result in each dimension. Tab.~\ref{tab:userstudy} shows that our method achieved superior performance across all dimensions, demonstrating its effectiveness in generating high-quality 3D characters.
\begin{table}
\centering
\resizebox{0.9\linewidth}{!}{
\begin{tabular}{l|cccc}
\toprule
Method & \begin{tabular}[c]{@{}c@{}}Overall\\Quality\end{tabular} & Fidelity & \begin{tabular}[c]{@{}c@{}}Geometric\\Quality\end{tabular} & \begin{tabular}[c]{@{}c@{}}Texture\\Quality\end{tabular} \\
\midrule
CharacterGen~\cite{peng2024charactergen} & 4.0\% & 4.5\% & 6.2\% & 4.2\% \\
InstantMesh~\cite{xu2024instantmesh} & 1.8\% & 2.0\% & 1.8\% & 2.5\% \\
Unique3D~\cite{wu2024unique3d} & 11.8\% & 18.3\% & 10.5\% & 17.0\% \\
Ours & \textbf{82.4\%} & \textbf{75.2\%} & \textbf{81.5\%} & \textbf{76.3\%} \\
\bottomrule
\end{tabular}
}
\caption{User study results comparing different methods across four dimensions. The values represent the percentage of times each method was selected as the best in the respective dimension.}
\label{tab:userstudy}
\vspace{-10pt}
\end{table}

\xhdr{Ablation Study}
We further conduct ablation studies on character decomposition and multi-layer refinement. Please see supplementary material (Sec.~A) for details.

\subsection{Applications}
\vspace{-10pt}
\begin{figure}[htb]
\centering
\includegraphics[width=0.95\linewidth]{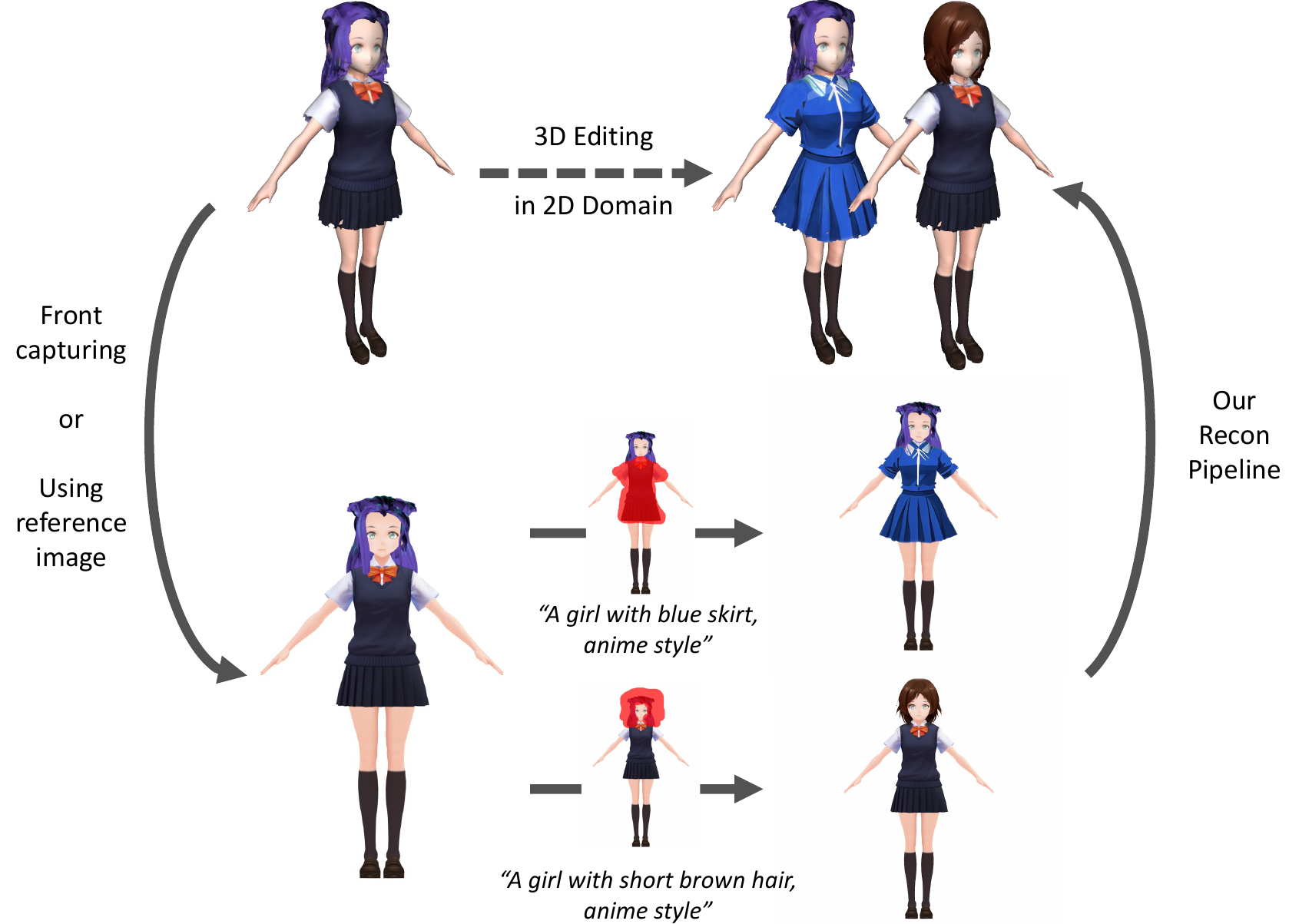}
\caption{Our pipeline enables diverse 3D editing using only text prompts, masks, and in-painting diffusion in the 2D domain.}
\label{fig:edit}
\vspace{-3pt}
\end{figure}
Our method's capability to generate decomposed characters in A-pose not only facilitates simpler rigging but also enables 3D editing through user-specified prompts and in-painting masks in the 2D domain. As illustrated in Fig.~\ref{fig:edit}, our framework allows for effortless character customization, including outfit changes and hairstyle modifications. The process requires minimal user input: the original reference image or a front-view capture, a roughly sketched in-painting mask, and a text prompt. Utilizing existing in-painting models (HD-Painter~\cite{manukyan2023hd} in this case), we achieve the desired 2D edits. Subsequently, these 2D modifications are translated into the 3D domain through our reconstruction and optimization pipeline. During the multi-layer refinement stage, the relevant mesh components are selectively replaced with the edited versions while preserving the unmodified parts. This bridges the gap between 2D image editing and 3D character customization and streamlines the 3D character modification process. We further provide animation comparisons in supplementary material (Sec.~B).

%% file: sec/5_conclusion.tex
\section{Conclusion}
\label{sec:conclusion}

In this paper, we introduce \ourmethod, an innovative pipeline for generating semantic decomposed high-quality 3D characters from single images. 
A novel semantic-aware large reconstruction model is proposed to jointly reconstruct geometry, color and semantics, together with an efficient 2D diffusion model and iterative multi-layer refinement module to enable the high-quality generation from arbitrary-posed single image.
\ourmethod\ achieves superior performance over existing baselines in terms of geometry, texture and decomposability, and offers ready-to-use, semantic-decomposed, customizable 3D characters, demonstrating significant potential for various applications.

\noindent\textbf{Acknowledgement.}\quad
This work was partially supported by Beijing Science and Technology plan project (Z231100005923029), Beijing Natural Science Foundation (L247007), and the Natural Science Foundation of China (62461160309, 62332019).

%% file: sec/X_suppl.tex
\clearpage
\appendix
\setcounter{page}{1}
\maketitlesupplementary

\section{Ablation Study}
\label{sec:ablation}

\begin{figure}[htb]
\centering
\includegraphics[width=1.0\linewidth]{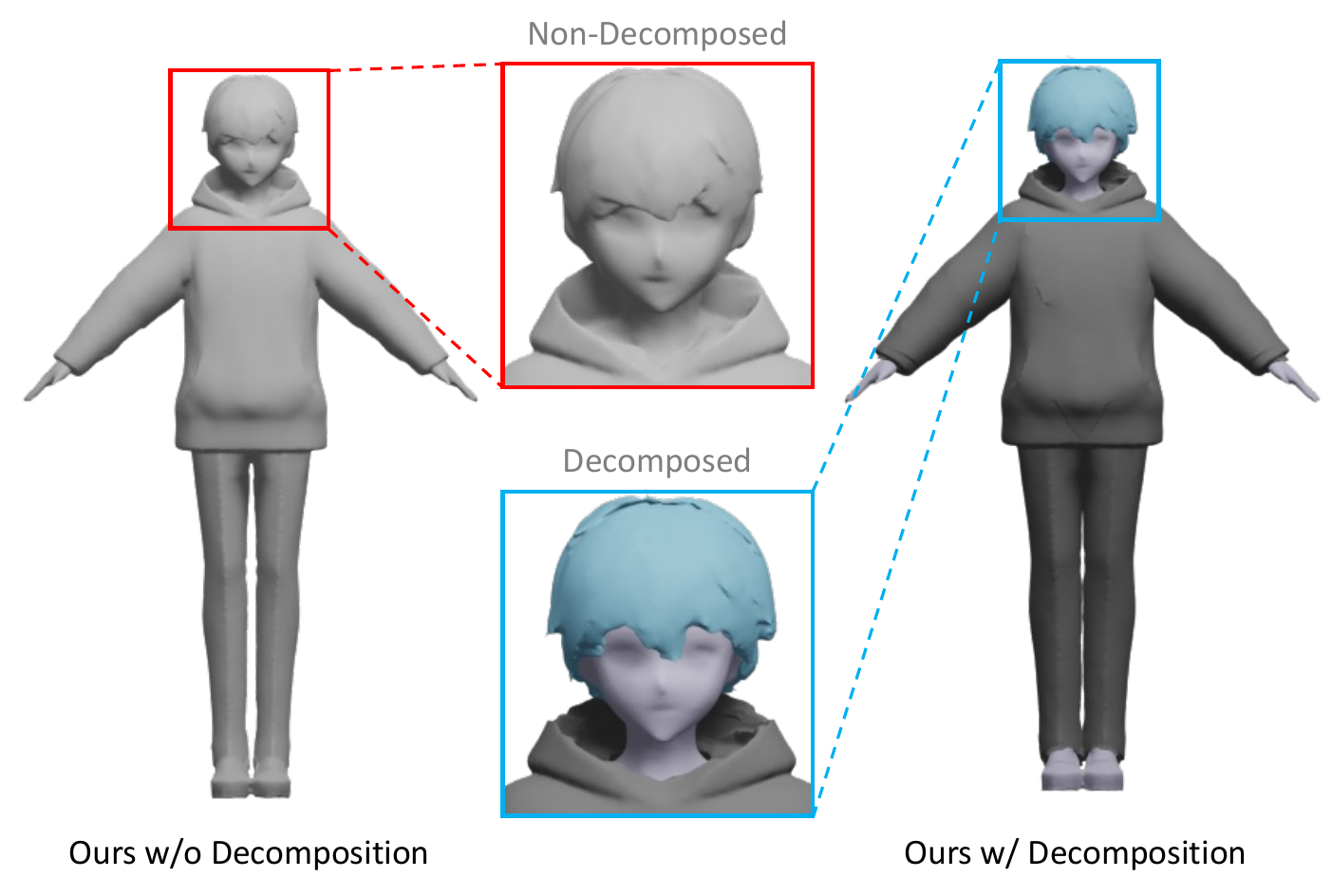}
\caption{Ablation study on character decomposition.}
\label{fig:abl_dec}
\end{figure}
\xhdr{Character Decomposition}
To demonstrate the decomposition capabilities of our core S-LRM and its impact on the results, we compared our method with a direct refinement approach that does not employ semantic decomposition. The visual comparison in Fig.~\ref{fig:abl_dec} reveals that without decomposition, the results exhibit a fusion of hair, clothing, and the base human model, significantly limiting their potential for downstream applications. In contrast, our method successfully separates these components while maintaining high mesh precision, showcasing the effectiveness of our semantic decomposing approach.

\begin{figure}[htb]
\centering
\includegraphics[width=1.0\linewidth]{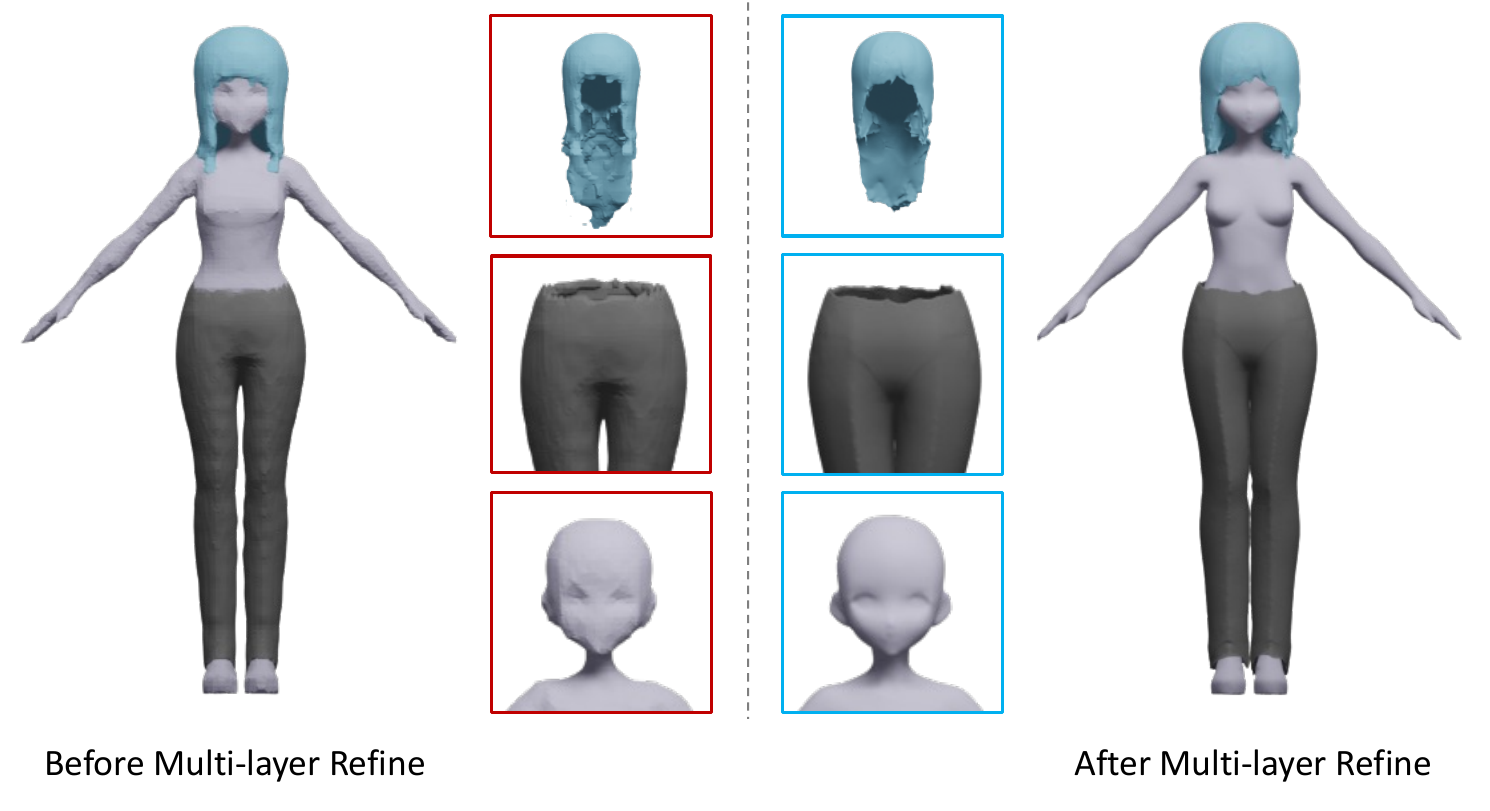}
\caption{Ablation study on multi-layer refinement. Zoom in for better details.}
\label{fig:abl_geo}
\end{figure}
\xhdr{Multi-layer Refinement}
We further illustrate the distinction between the direct output of our S-LRM and the results after multi-layer refinement in Fig.~\ref{fig:abl_geo}. The pre-optimization results demonstrate that our S-LRM successfully decomposes various mesh components with correct geometry and shape, validating the capabilities of our S-LRM. However, the precision is limited due to the inherent characteristics of FlexiCubes~\cite{shen2023flexible} and memory constraints. Post-refinement, we observe a substantial enhancement in precision while maintaining the overall structure. This improvement underscores the effectiveness of our multi-layer refinement process in preserving the structure of the decomposed components while significantly elevating the geometric accuracy and overall quality of the reconstructed character.

\section{More Applications}
\label{sec:moreapp}
\begin{figure}[t]
\centering
\includegraphics[width=1.0\linewidth]{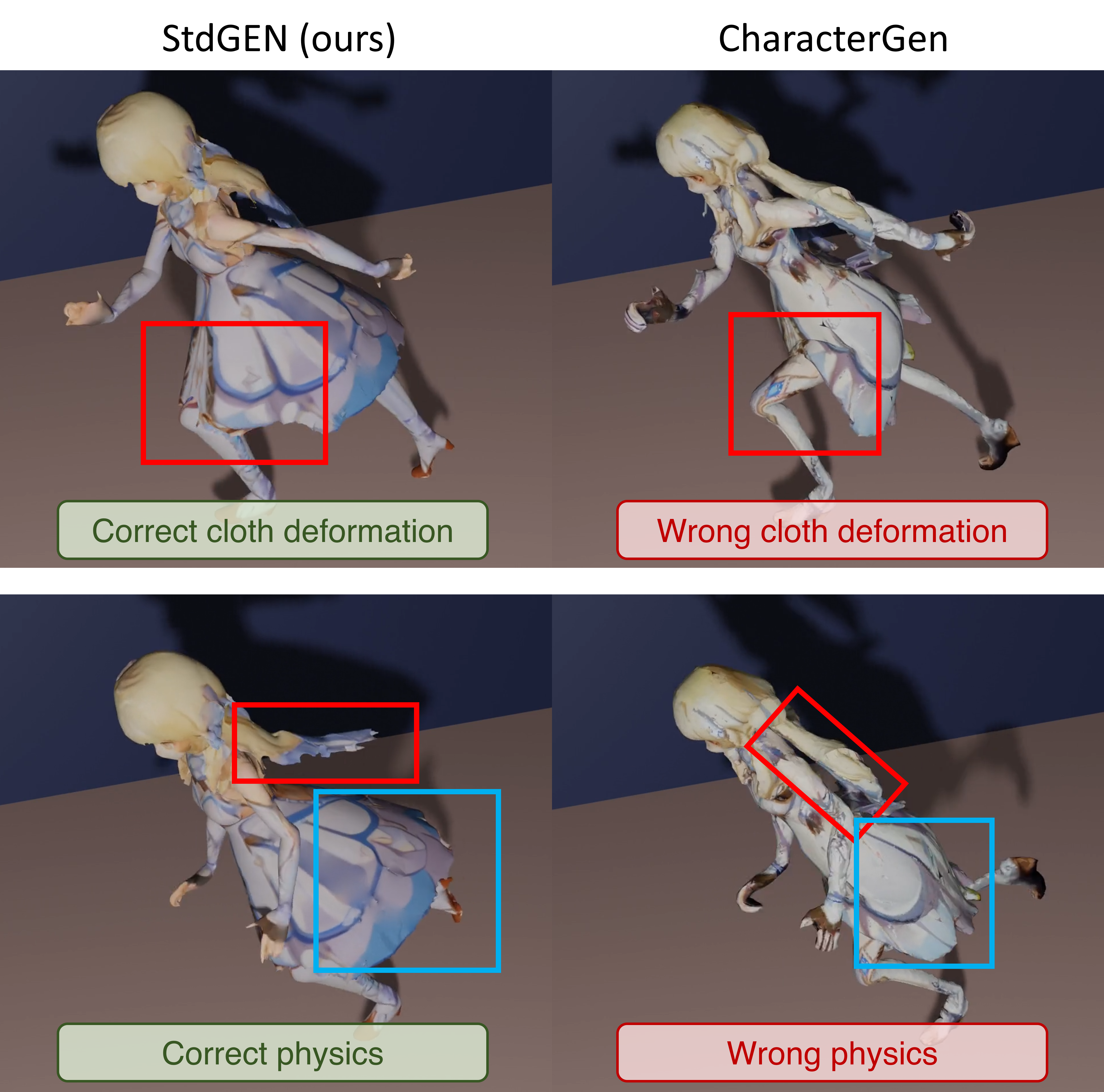}
\caption{Rigging and animation comparisons on 3D character generation. Our method demonstrates superior performance in human perception and physical characteristics.}
\label{fig:anim}
\end{figure}
Compared to other 3D character generation methods, our decomposed generation in A-pose is more suitable for downstream animation and 3D applications. In Fig.~\ref{fig:anim}, we rig the 3D character generated by our approach and by CharacterGen~\cite{peng2024charactergen} for comparison. Without decomposition, the hair and clothing stick together and are attached to the base human model. In contrast, our approach maintains separated parts, aligning more closely with natural perception. Additionally, the non-decomposed nature leads to inaccurate deformations and physical characteristics during movement, which our method effectively avoids.

\section{Time Breakdown Analysis}
\begin{table}[h]
\centering
\vspace{-10pt}
\begin{tabular}{>{\raggedright}p{4.5cm} >{\centering\arraybackslash}p{2cm}}
\toprule
\textbf{Process} & \textbf{Time (s)} \\
\midrule
\textbf{Canonicalization Diffusion} & 7 \\
\textbf{Multi-view Diffusion} & 29 \\
\textbf{S-LRM} & 12 \\
\textbf{Refinement} & \\
\hspace{3mm} - Single-layer Settings & 18 \\
\hspace{3mm} - Multi-layer Settings & 117 \\
\bottomrule
\end{tabular}
\caption{Time breakdown for each processing step.}
\label{tab:time}
\end{table}
In Tab.~\ref{tab:time}, we present the time breakdown of different components in our method. Creating a single-layered 3D character takes only about 1 minute while generating a decomposed, multi-layered 3D character requires less than 3 minutes. Our S-LRM is relatively efficient, with minimal additional time and memory overhead compared to InstantMesh's LRM.

\section{More Quantitative Results}
\xhdr{3D Semantic Metrics}
To demonstrate our 3D semantic decomposition capability, we extracted separate meshes for three semantic categories (hair, cloth, and base human model) and rendered masks from eight different views for comparison with ground truth. 
In the arbitrary pose setting, we achieved IoU scores of 0.73 for hair, 0.86 for cloth, and 0.88 for the base human model. 
These results demonstrate effective semantic decomposition, particularly considering the significant challenges in single-image-based 3D reconstruction, such as spatial ambiguity and occlusion.

\xhdr{3D Geometric Metrics}
Additional comparisons in the arbitrary pose setting are presented in Tab.~\ref{tab:re-3d}, where our method demonstrates superior performance on both the Chamfer distance and the F-score, showing a more precise prediction of 3D geometry.

\begin{table}[htb]
\centering
\resizebox{1.0\linewidth}{!}{
\begin{tabular}{l|c|c|c}
\toprule
Metric & Unique3D & CharacterGen & Ours \\
\midrule
Chamfer Distance $\downarrow$ & 0.109 & 0.035 & \textbf{0.023} \\
F-Score $\uparrow$ & 0.137 & 0.465 & \textbf{0.664} \\
\bottomrule
\end{tabular}
}
\caption{3D metric comparison (0.01 for f-score threshold).}
\label{tab:re-3d}
\end{table}

\xhdr{Ablation Study on Refinement Stage}
We conduct experiments without refinement in Tab.~\ref{tab:re-3dgeo} under arbitrary pose setting, where we only apply color back-projection to the mesh generated by S-LRM.
The results indicate that our method outperforms CharacterGen and Unique3D even without refinement, demonstrating that our S-LRM is effective even without the refinement step.

\begin{table}[htb]
\centering
\resizebox{1.0\linewidth}{!}{
\begin{tabular}{lcccc}
\toprule
Method & SSIM $\uparrow$ & LPIPS $\downarrow$ & FID $\downarrow$ & CLIP Similarity $\uparrow$ \\
\midrule

Unique3D & 0.856	& 0.190	& 0.042 &	0.903 \\
CharacterGen & 0.869	& 0.134	& 0.119 &	0.901 \\
Ours (w/o refine) & 0.912 & 0.094 & 0.026 & 0.933 \\
Ours     & \textbf{0.916}	& \textbf{0.084}	& \textbf{0.011} &	\textbf{0.936} \\

\bottomrule
\end{tabular}
}
\vspace{-3pt}
\caption{Ablation study on refinement stage.}
\vspace{-12pt}
\label{tab:re-3dgeo}
\end{table}

\section{Dataset Construction}
\label{sec:dataset}

\xhdr{Semantic Definition}
Unlike general 3D models, 3D character modeling typically involves multiple components rather than a single entity. This segmentation is essential for downstream applications such as rigging and physical simulation, which often require the manipulation of distinct parts. Conventional reconstruction models only capture the character's surface appearance, lacking internal information and the ability to decompose the model, which limitation severely restricts subsequent applications.
After considering both practical applications and data composition, we have categorized character composition into three semantic categories: base minimal-clothed human model, clothing, and hair (specifically, shoes and underwear are classified as part of the base human model, considering downstream applications and data characteristics). By incorporating these semantic categories into our reconstruction process, we aim to produce 3D character models that are not only visually accurate but also functionally versatile for various uses in 3D game and animation pipelines.
Note that our method supports the learning and extracting of an arbitrary number of semantic categories.

\xhdr{Data Cleaning}
We begin by filtering out data that cannot be layered according to semantic structure. Using multiple prompts combined with ImageReward~\cite{xu2024imagereward}, we remove low-quality or malicious data with low scores. Additionally, we identify instances where semantic information predictions are inaccurate, then manually review and remove data that appears semantically incorrect to human perception. Since the base human model in the original dataset can occasionally contain defects, we apply a connectivity check on the front rendering of every base human model. Examples lacking connectivity are only used to supervise either the complete model or the base human model with clothing, but not the base human model alone.

\xhdr{Rendering Settings}
Our rendering process goes beyond standard image generation, incorporating depth, normal, and semantic maps. 3D Character models are adjusted to an A-pose configuration, with arms rotated 45 degrees downward from the horizontal position. To facilitate multi-layered reconstruction supervision, we rendered the complete model and two additional configurations: the base minimal-clothed human model alone, and the base model with clothing. The rendering includes eight views at 45-degree azimuth intervals with zero elevation, supplemented by top-down and bottom-up views. We enriched the dataset with five close-up facial views and 20 random viewpoints. To enhance the training of our diffusion model, we implemented data augmentation on varying arm angles.

We use the orthographic camera for all renderings. For non-close-up views of the character, after normalizing the character to fit within a unit cube, we set the ortho scale to 1.2. For close-up views of the face, we locate the 3D center position and bounding box of the facial semantics, aligning the camera’s center projection with the 3D center of the face and setting the ortho scale to 1.2 times the bounding box size. We render five facial close-up views at elevation = 0° and azimuth angles of $\{-90^\circ, -45^\circ, 0^\circ, 45^\circ, 90^\circ\}$. For inputs to the canonicalization diffusion model, we add outline and shading with a 50\% probability. Rim lighting, shading, and outlines (except on the face) are consistently removed to supervise diffusion and S-LRM outputs. Semantic maps are rendered by modifying non-transparent regions of the texture map assigned to specific semantic parts.

\xhdr{Multi-layer Settings}
We provide three different rendering levels: the complete model, the base human model only, and the base human model with clothes, each generated by selectively removing specific semantic elements. For supervising S-LRM, these correspond to (1) no semantic masking, (2) masking of hair and clothing, and (3) masking of hair only. By mixing these levels of 2D supervision, we can train S-LRM to reconstruct multi-layered density, color, and semantic information automatically.

\section{Details of Loss Functions}
In this section, we provide a detailed description of each loss component in our framework.
$\mathcal{L}_{\text{mse}}$ is the commonly used mean squared error loss defined as:
\begin{align}
     \mathcal{L}_{\text{mse}} &= \sum_k \left\|\hat{I}_k-I_k^{gt}\right\|_2^2
\end{align}
where $\hat{I}_k$, $I_k^{gt}$ denotes the $k$-th view of rendered images and ground-truth images, respectively.

$\mathcal{L}_{\text{lpips}}$ is the perceptual loss defined as
\begin{align}
    \mathcal{L}_{\text{lpips}} &= \sum_k \tau\left(\phi(\hat{I}_k), \phi(I_k^{gt})\right)
\end{align}
where $\phi$ is the VGG feature extractor, $\tau$ transforms deep embedding to a scalar LPIPS score.

$\mathcal{L}_{\text{mask}}$ is the mask loss defined as
\begin{align}
    \mathcal{L}_{\text{mask}} &= \sum_k \left\|\hat{M}_k-M_k^{gt}\right\|_2^2
\end{align}
where $\hat{M}_k$ and $M_k^{gt}$ denote the rendered non-transparent mask, and ground-truth masks, respectively.

The deviation loss \(\mathcal{L}_{\text{dev}}\) penalizes the Euclidean distances between each dual vertex \(v\) and the edge crossings \(u_e \in \mathcal{N}_v\) that bound its primal face, encourages vertices to center within their cells and allowing flexibility for connectivity adaptation:
\begin{align}
    \mathcal{L}_{\text{dev}} = \sum_{v\in V} \text{MAD}[\{ |v-u_e|_2 \ : \ u_e\in\mathcal{N}_v \}]
\end{align}
where $|\cdot|_2$ is the Euclidean distance, \(\text{MAD}(Y) = \frac{1}{|Y|}\sum_{y \in Y}|y - \text{mean}(Y)|\) denotes the mean absolute deviation. We use the same approach as the implementation of InstantMesh~\cite{xu2024instantmesh}, applying L2 regularization with a weight of 0.1 to the FlexiCubes weights.

During S-LRM training, we specifically incorporated facial semantics as an additional component to enhance the training process and facilitate potential applications. In subsequent stages, facial semantics were treated as an integral part of the base human model. For the semantic cross-entropy loss $\mathcal{L}_{\text{sem}}$, we empirically assigned weights to four semantic categories - hair (1.255), face (1.758), base human model (0.913), and cloth (0.650) - based on their respective rendering proportions in the dataset to optimize semantic learning.

\section{Implementation Details}
We divide our Anime3D++ dataset into a training and testing set in a 99:1 ratio. 
We first train the canonicalization diffusion model at a 512 resolution with a learning rate of 5e-5, then reduce it to 1e-5 as we progressively increase the resolution to 768 and 1024. Similarly, the multi-view diffusion model is trained at a constant learning rate of 5e-5 while scaling from 512 to 768 and finally to 1024 resolution.
We use LoRA with a 128-rank for S-LRM, a learning rate of 4e-5, and three supervision stages with rendering resolutions of 192, 144, and 512. The loss function parameters are set as $\lambda_{\text{lpips}}, \lambda_{\text{mask}}, \lambda_{\text{sem}}, \lambda_{\text{depth}}, \lambda_{\text{normal}}, \lambda_{\text{dev}} = 2.0,1.0,1.0, 0.5, 0.2, 0.5$.
For multi-layer refinement, we set $\lambda_{\text{mask}}', \lambda_{\text{normal}}', \lambda_{\text{col}} = 1.0, 1.0, 100.0$, and we further extract the coarse hair mask, applying additional normal and mask loss for hair refinement with a weight of 1 and 10.

\xhdr{Detailed Structure of S-LRM}
Following InstantMesh~\cite{xu2024instantmesh}, our S-LRM adopts 6 RGB images generated by multi-view diffusion in a resolution of $320\times 320$ as model input.
In the training stage 3 (training on meshes with multi-layer semantics), we set the sampling grid size for FlexiCubes extraction to $100\times 100\times 150$, with dimensions scaled to $0.7\times 0.7\times 1.05$ of the triplane size, and the centers of both are aligned. We integrate the LoRA~\cite{hu2021lora} structure into the S-LRM transformer, modifying both the self-attention and cross-attention modules.
For self-attention, where $q$, $k$, and $v$ values are produced by shared linear layers, we substitute all input and output linear layers with LoRA structures. In cross-attention, where $q$, $k$, and $v$ are produced through separate linear layers, we replace the linear layers for $q$, $k$, $v$, and the outputs with LoRA structures. The specifics are as follows:
\begin{align}
    h^i &= W_0^i + \Delta W_{tp}^ix = W_0^ix + B_{tp}^iA_{tp}^ix
\end{align}
Here, \( i \) denotes the $i$-th transformer layer. In self-attention, \( tp \) represents the linear projection for inputs and outputs, while in cross-attention, \( tp \) denotes the linear projections for $q$, $k$, $v$, and outputs.
During the training process, the DINO~\cite{caron2021emerging} encoder is kept frozen while the feature decoder (including color/density decoder and semantic decoder) remains trainable. In the triplane transformer, the positional embeddings, de-convolution layers, and all LoRA layers are set as learnable parameters, while all other layers are frozen.

\xhdr{Detailed Settings of Canonicalization Diffusion}
Our canonicalization diffusion model comprises a U-Net and a ReferenceNet with an identical architecture, both networks are initialized with the weights from Stable Diffusion 2.1. The U-Net takes the CLIP-encoded features of the reference image as input for the encoder-hidden states. ReferenceNet, on the other hand, receives the image latents of the reference image (encoded by a VAE without added noise) as input, along with the features of a fixed text prompt, “high quality, best quality,” encoded by CLIP~\cite{radford2021learning}, which are fed into the encoder-hidden states. A cross-attention operation is applied for each corresponding layer pair in the U-Net and ReferenceNet, using the current U-Net layer as the query and the corresponding ReferenceNet layer as the key and value. This cross-attention mechanism transfers the detailed information of the reference image into the U-Net.

\xhdr{Detailed Settings of Multi-View Diffusion}
Building upon Era3D's~\cite{li2024era3d} multi-view model, we start training from its inherited weights. We concatenate the noisy VAE latent and reference image VAE latent as input to the U-Net. For each view's color and normal output, we specify fixed prompts ("a rendering image of 3D models, \{view\} view, \{color/normal\} map"), which are encoded through CLIP and fed into the U-Net's encoder hidden states. For U-Net's class labels, we reserve the first 1024 dimensions for the CLIP embedding of the reference image, and replace the noise level embedding in the latter 1024 dimensions with a level switcher. This level switcher uses different one-hot vectors for three distinct rendering levels to support the specific semantic combinations in the diffusion output, serving as supervision signals for multi-layer refinement.
Since the previous canonicalization diffusion step already ensures that the output A-pose character reference image has elevation=0 and is orthographic, we do not employ the regression loss from Era3D. Additionally, we fix the noise level of the image VAE latent to 0 to achieve optimal fidelity.

\xhdr{Details of Color Back-projection}
We employ a multi-view projection method similar to Unique3D~\cite{wu2024unique3d} to back-project the texture onto the 3D character model. For each vertex $v$ that is visible in at least one view, we calculate its final color $C(v)$ in the 3D mesh using the following formula:
\begin{align}
    C(v)=\sum_{i\in I}\frac{w_i(\mathbf{n}_v\cdot\mathbf{d}_i)^2c_{v,i}}{w_i(\mathbf{n}_v\cdot\mathbf{d}_i)^2}
\end{align}
where $c_{v,i}$ represents the color corresponding to $v$ in the $i$-th view texture; $\mathbf{n}_v$ and $\mathbf{n}_i$ is the vertex normal of $v$ and the view direction of the $i$-th view respectively; $w_i$ is the projection weight of the $i$-th view, and $I$ is the set of views where $v$ is visible.
In practice, we assign $w_i$ values of $\{2.0, 0.5, 0.0, 1.0, 0.0, 0.5\}$ for views at azimuths of $\{0^\circ, 45^\circ, 90^\circ, 180^\circ, 270^\circ, 315^\circ\}$ respectively. 
For vertices that are not visible in any view, we treat the 3D mesh as a graph composed of vertices and edges, and iteratively perform convolution and mean pooling to transfer colors from vertices with determined colors to those without, until all vertices obtain their colors.

\xhdr{Pre- and Post-dilation of Mesh}
To better solve the problematic intersections among meshes in the multi-layer refinement stage, we introduce a "dilation" process applied both before and after optimization. This process constructs an approximate "flow field" based on the original positions of the inner and outer layer meshes.

For each vertex on the outside mesh, we utilize a kd-tree to query its nearest vertex neighbors of the fixed inside mesh. The movement range is then smoothly weighted based on the exponential inverse distance from these neighbors, with distant points remaining stationary. This approach creates a "dilation" effect, ensuring that when inner layers are moved outward to resolve intersections, the outer layers follow suit in a natural, gradual manner.

\section{Discussions}

\xhdr{Comparison with other decomposition methods}
We note that some works have also applied the concept of decomposition, while they differ from our method in problem definition and scope. GALA~\cite{kim2024gala} and TELA~\cite{dong2024tela} use real-world 3D mesh scans and text as input respectively, employing Score Distillation Sampling (SDS)~\cite{poole2022dreamfusiontextto3dusing2d} with class-specific text prompts and pose control for layered 3D avatar generation. Frankenstein~\cite{yan2024frankenstein} takes a textureless, 2D semantic layout as input and generates semantic-decomposed 3D meshes (also textureless) through triplane diffusion. In contrast, our method accepts RGB reference images of arbitrary characters and generates 3D character meshes that faithfully preserve the reference texture while enabling semantic decomposition in a feed-forward manner.

Regarding specific decomposition techniques, GALA and TELA use SDS and different prompts to optimize both individual parts and the whole iteratively, typically requiring hours or more for a single case;
Frankenstein outputs separate SDFs for each semantic class, training and inferring on datasets with specific semantics, while our method treats geometry and semantics information independently, enabling greater compatibility and scalability. Our method can extract equivalent surfaces by specifying any combination of semantics, while maintaining compatibility with general datasets like Objaverse~\cite{deitke2023objaversexluniverse10m3d} and preserving LRM's general performance. It also has the potential to achieve semantic decomposition for multiple data types through a single LRM paired with multiple semantic decoders. Moreover, when adding new semantic classes, our approach can inherit geometric priors, making fine-tuning more efficient.

\begin{figure}[htb]
\centering
\includegraphics[width=1.0\linewidth]{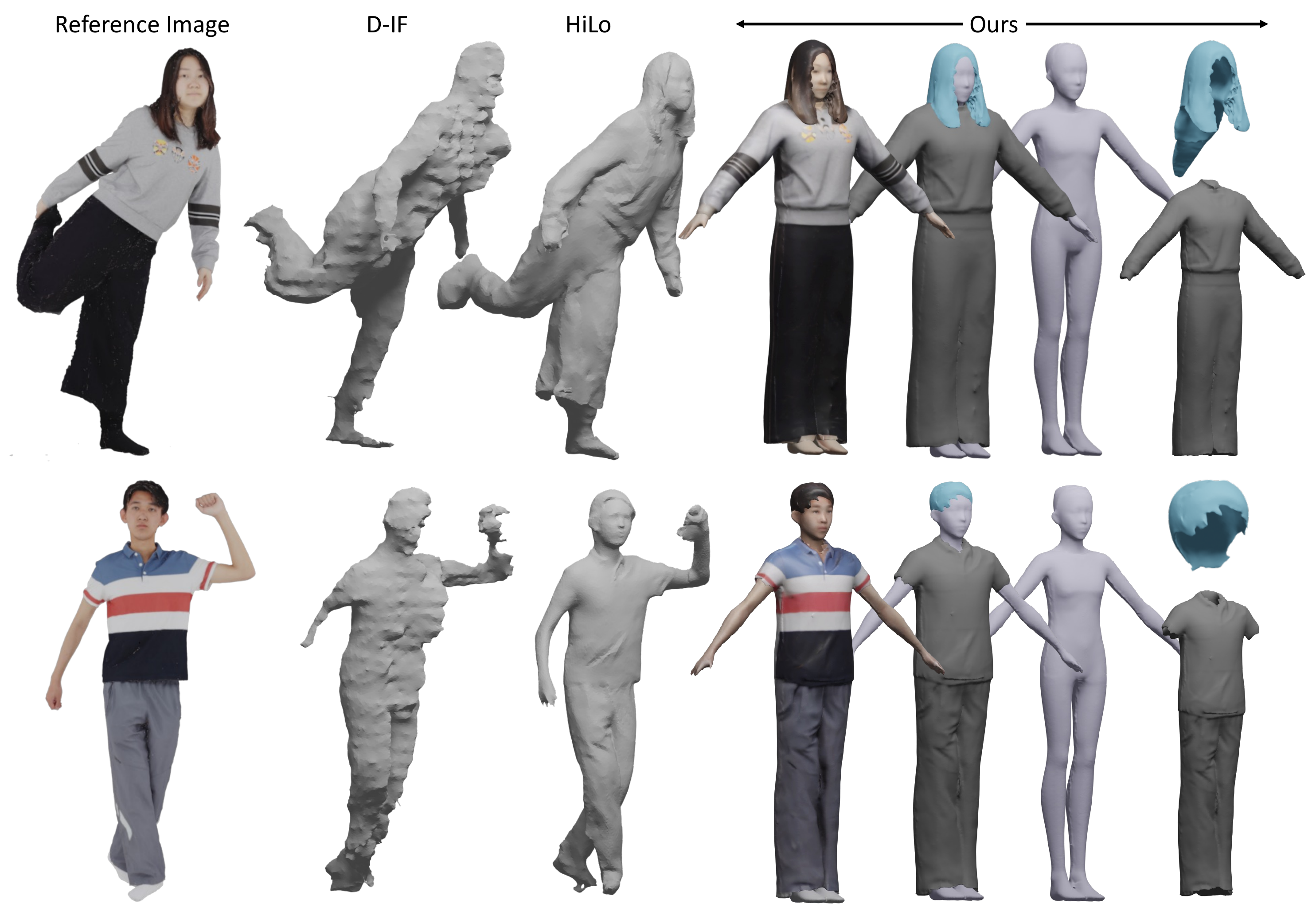}
\caption{Comparison on THuman 2.0 dataset.}
\label{fig:re-human}
\end{figure}

\begin{figure}[htb]
\centering
\vspace{-15pt}
\includegraphics[width=1\linewidth]{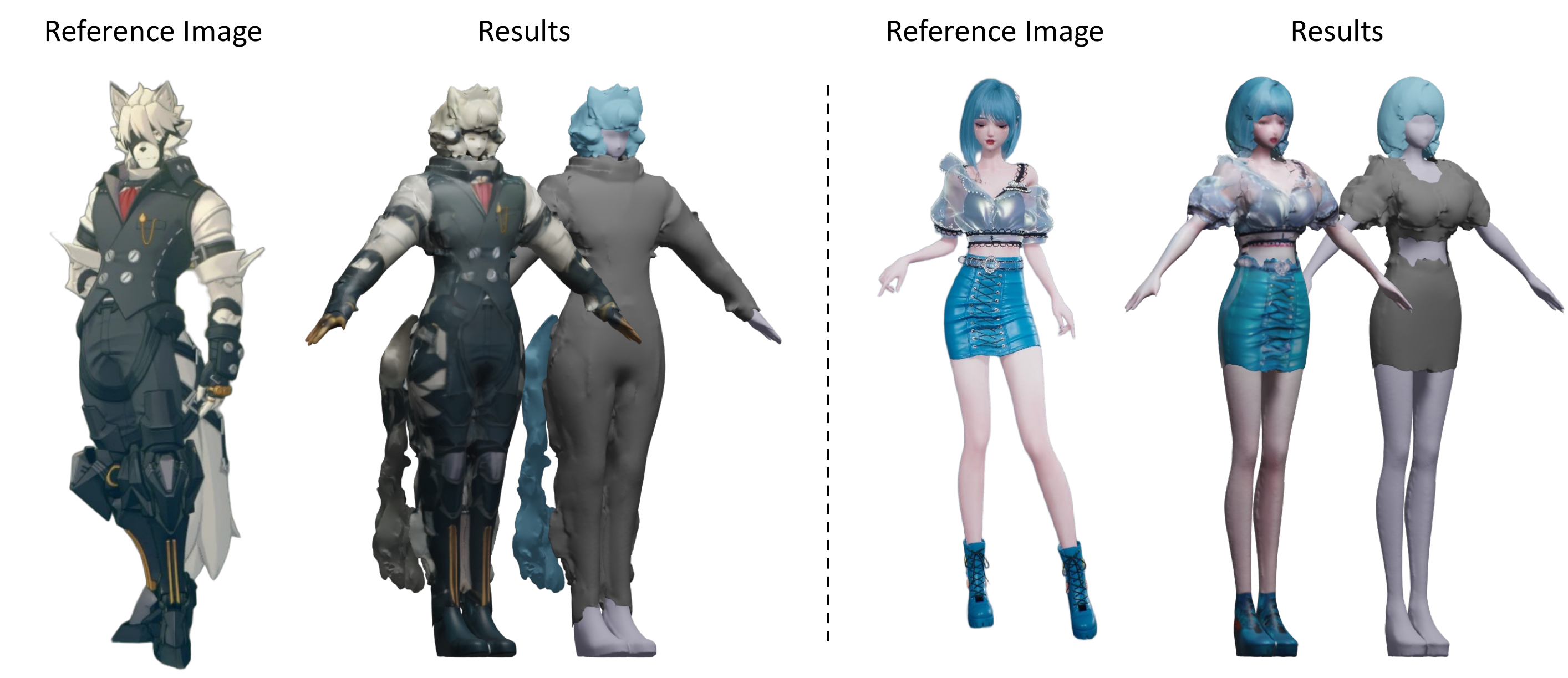}
\caption{Our result on furry and 2.5D style images.}
\label{fig:re-otherstyle}
\end{figure}

\begin{figure}[t!]
\centering
\includegraphics[width=1.0\linewidth]{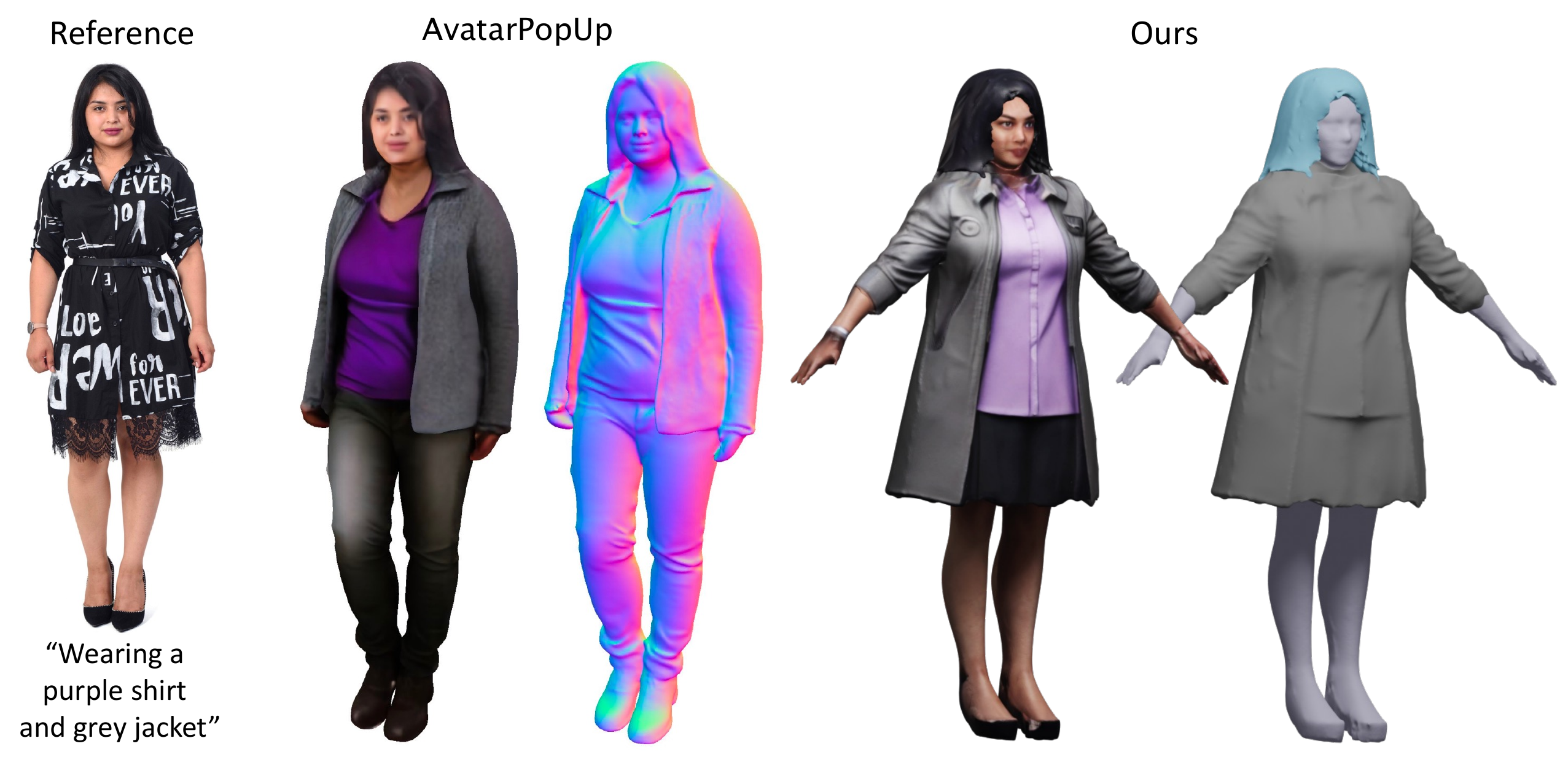}
\caption{3D editing comparison with AvatarPopUp.}
\label{fig:re-edit}
\vspace{-10pt}
\end{figure}

\xhdr{Non-anime style results}
Our method demonstrates generalization across diverse character types, as shown in Figs.~\ref{fig:re-human} and \ref{fig:re-otherstyle}.
For realistic style, we compare with D-IF~\cite{yang2023d} and HiLo~\cite{yang2024hilo} on THuman 2.0 dataset~\cite{yu2021function4d}.
While the results show slight stylistic bias inherently from the Anime3D++ training data (e.g., slim faces), our method is general with robustness, canonicalization, and decomposition capabilities even without real-human training data.
To further show our method's 3D editing and decomposing ability, we also directly compare the editing case in AvatarPopUp~\cite{kolotouros2024instant} (Fig.~\ref{fig:re-edit}). Our approach shows similar effectiveness on real-human examples as AvatarPopUp and offers semantic decomposition and style flexibility capabilities.

\xhdr{Semantic definition and possible improvements}
Our Anime3D++ dataset adopts the VRoid-Hub data standard, which is designed to align with VR/game requirements, particularly for animation and collision detection. 
Following this standard, close-fitting garments are classified as part of base human model, while outerwear~(e.g., pants, skirts, hoodies with long sleeves)~is categorized as clothing. 
Although our results currently support relatively few semantic categories due to the limitations in datasets, our method is general and the results demonstrate the feasibility of semantic awareness generation. In future work, our framework can be easily extended to support fine-grained semantic decomposition by incorporating Segment Anything Model (SAM)~\cite{kirillov2023segment} to generate detailed semantic labels for S-LRM training.

\begin{figure*}[htb]
\centering
\includegraphics[width=0.9\linewidth]{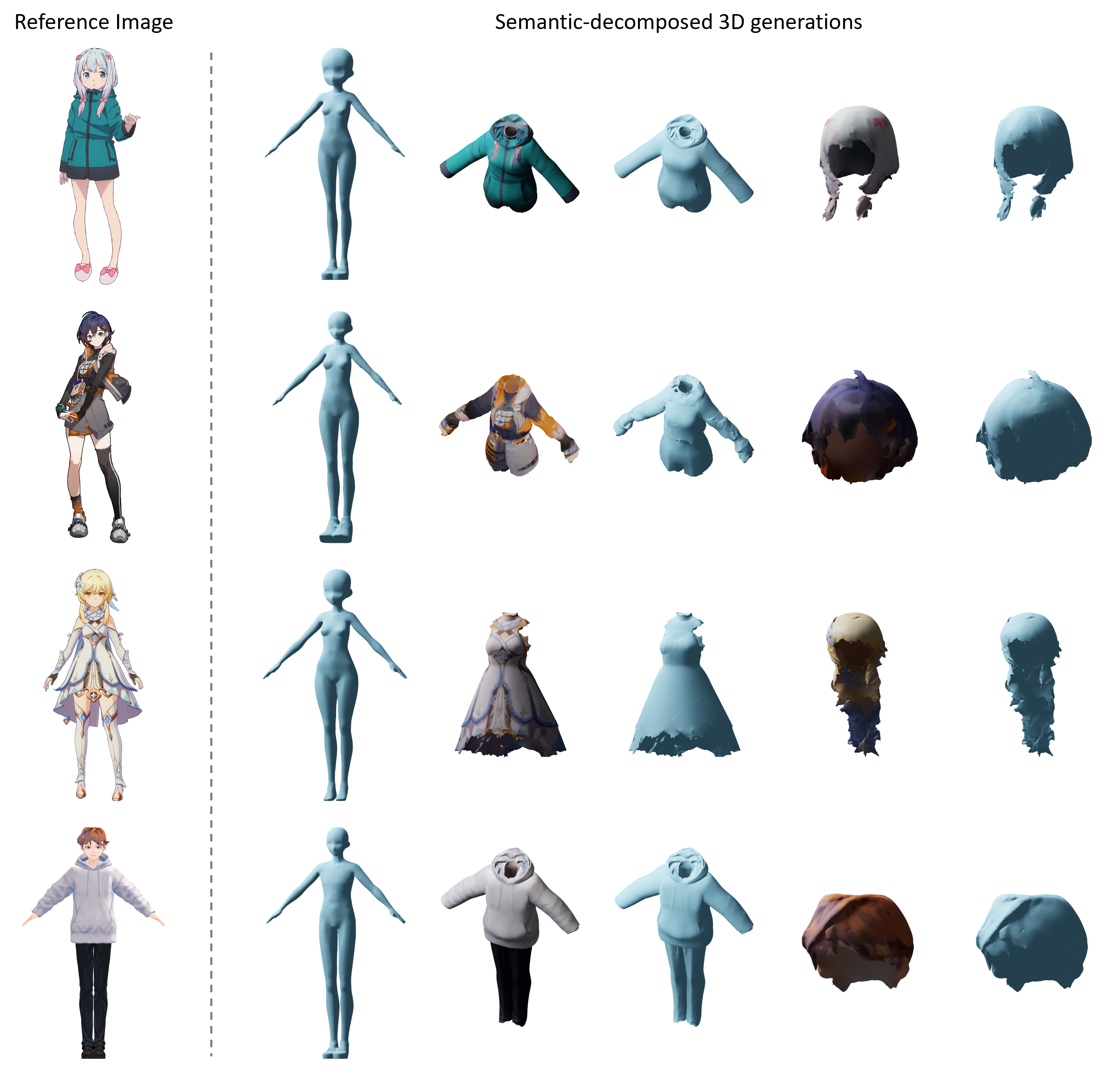}
\vspace{-10pt}
\caption{More visualizations on semantic-decomposed 3D generations.}
\label{fig:moredecomp}
\vspace{-7pt}
\end{figure*}

\begin{figure*}[htbp]
\centering
\includegraphics[width=0.95\linewidth]{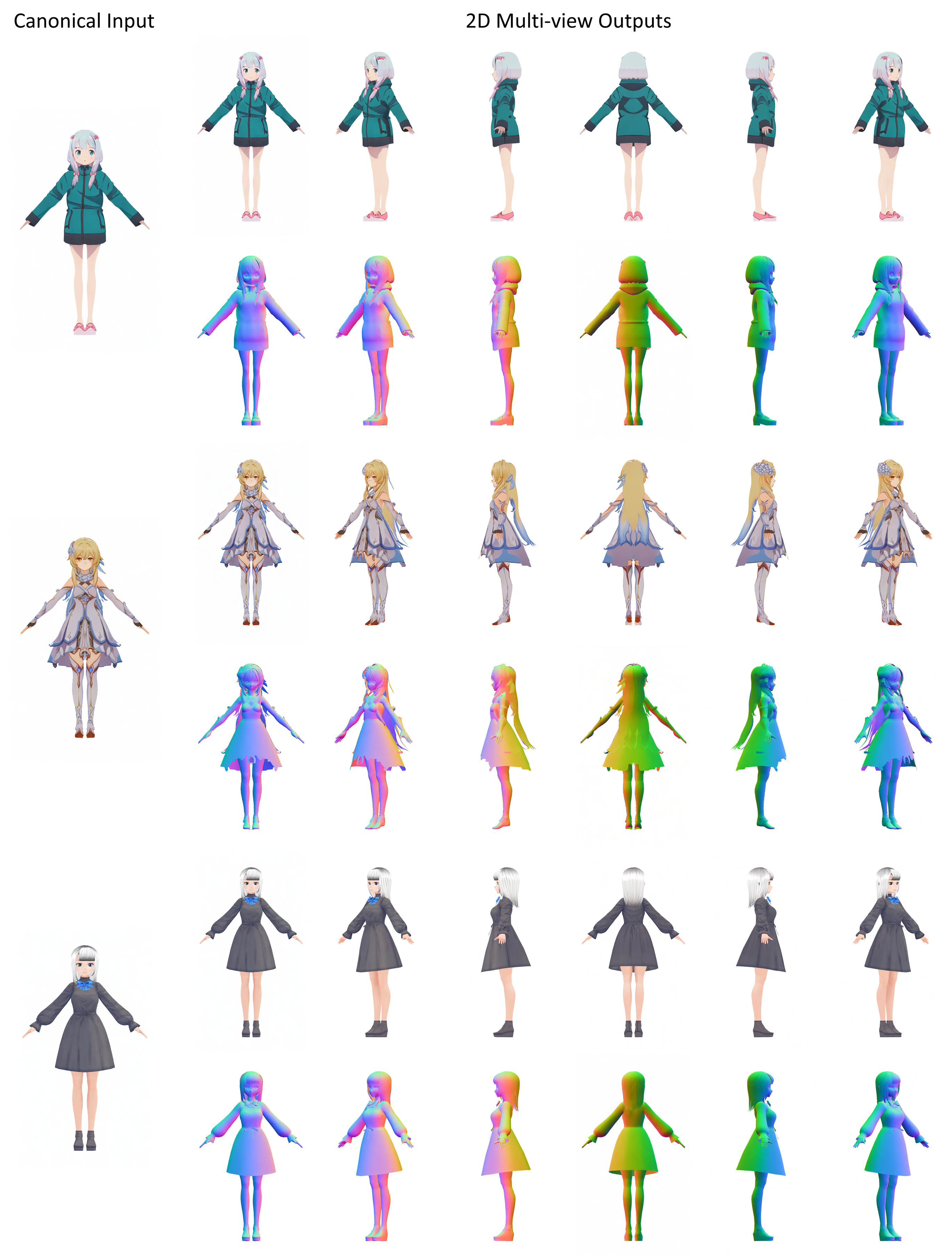}
\caption{Visualizations of the 2D multi-view diffusion model results.}
\label{fig:diff2}
\end{figure*}

\begin{figure*}[htbp]
\centering
\includegraphics[width=0.94\linewidth]{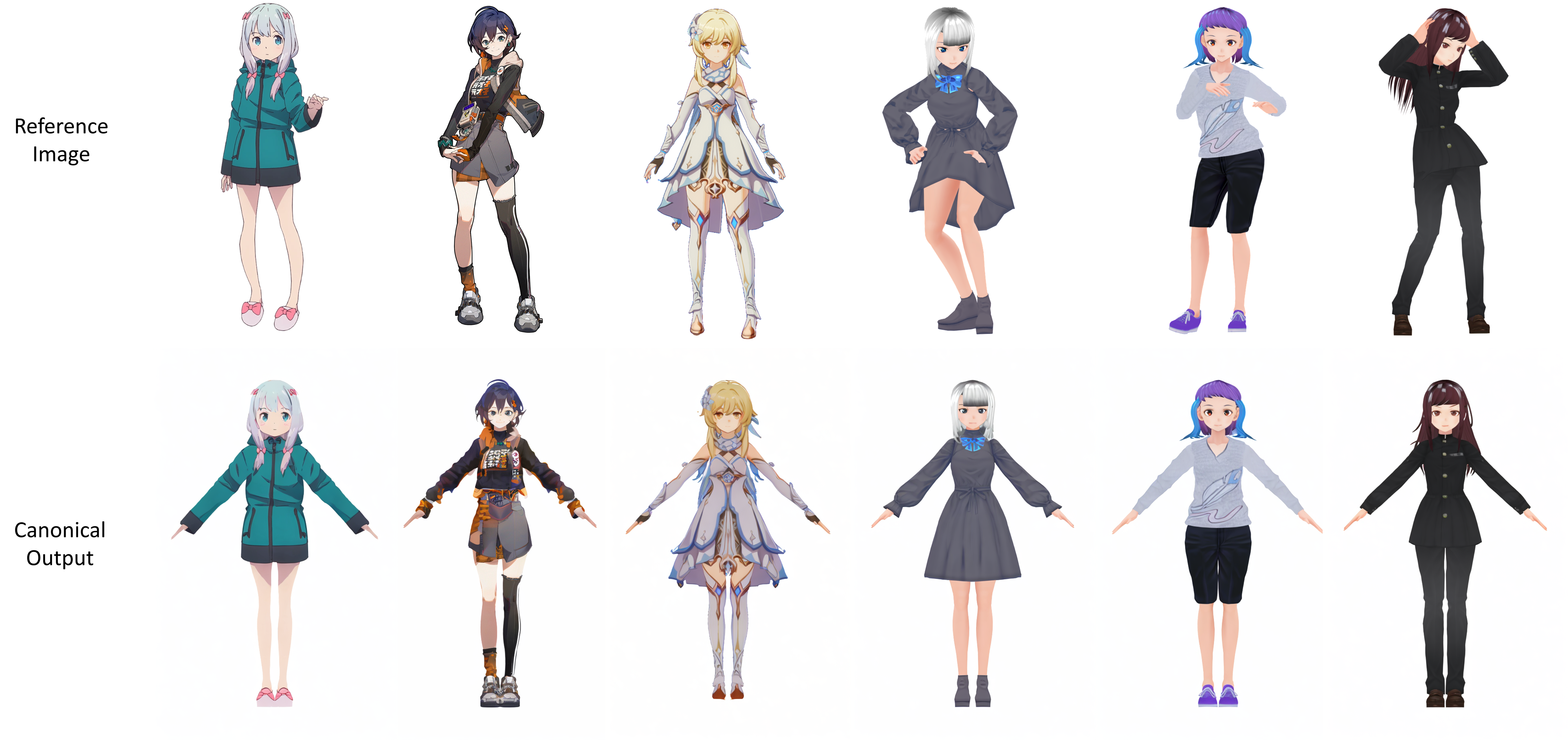}
\caption{Visualizations of the 2D canonicalization diffusion model results.}
\label{fig:diff1}
\end{figure*}

\begin{figure*}[htbp]
\centering
\includegraphics[width=0.94\linewidth]{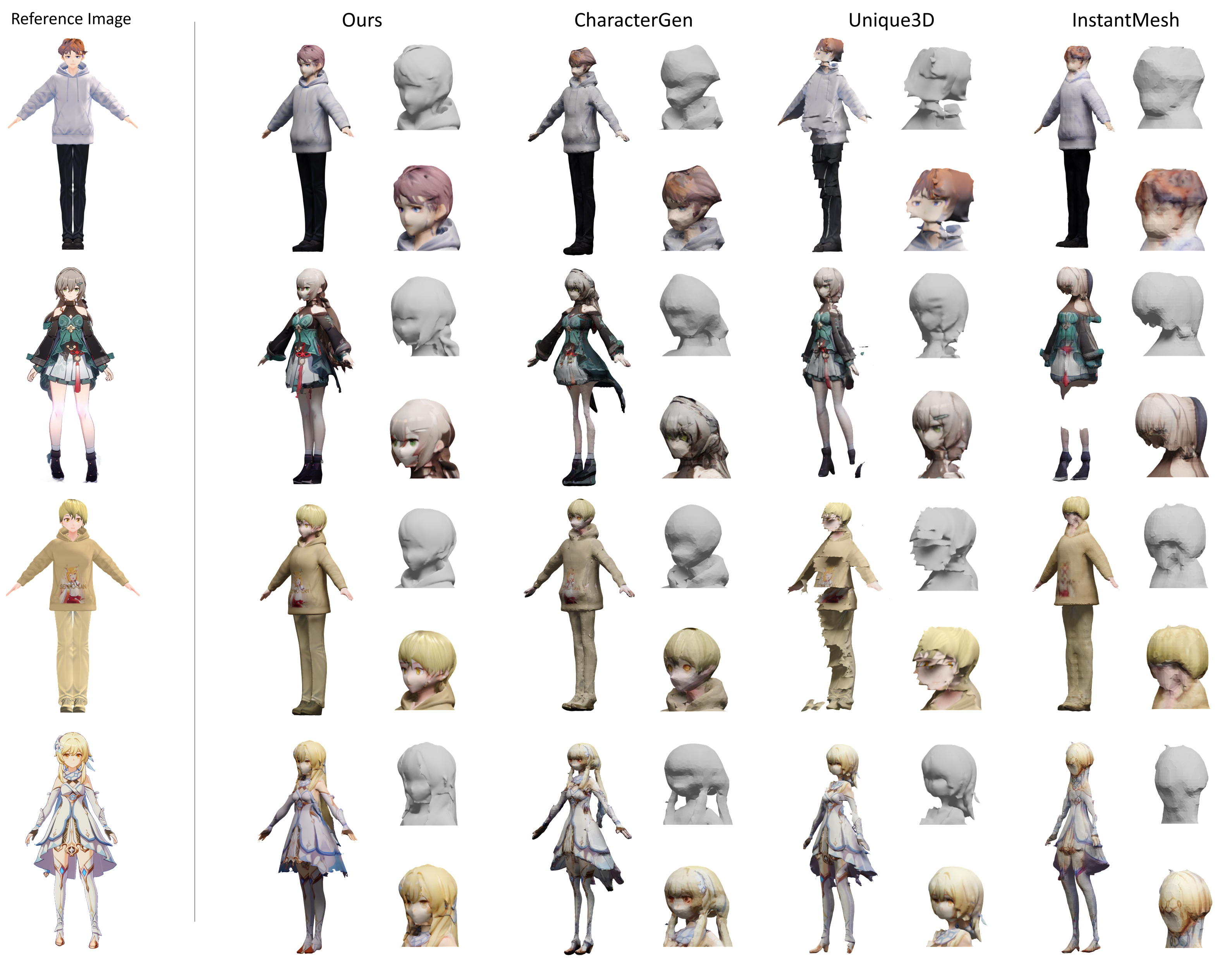}
\caption{More qualitative comparisons of 3D character generations (\#1).}
\label{fig:morequal}
\end{figure*}

\begin{figure*}[htbp]
\centering
\includegraphics[width=1.0\linewidth]{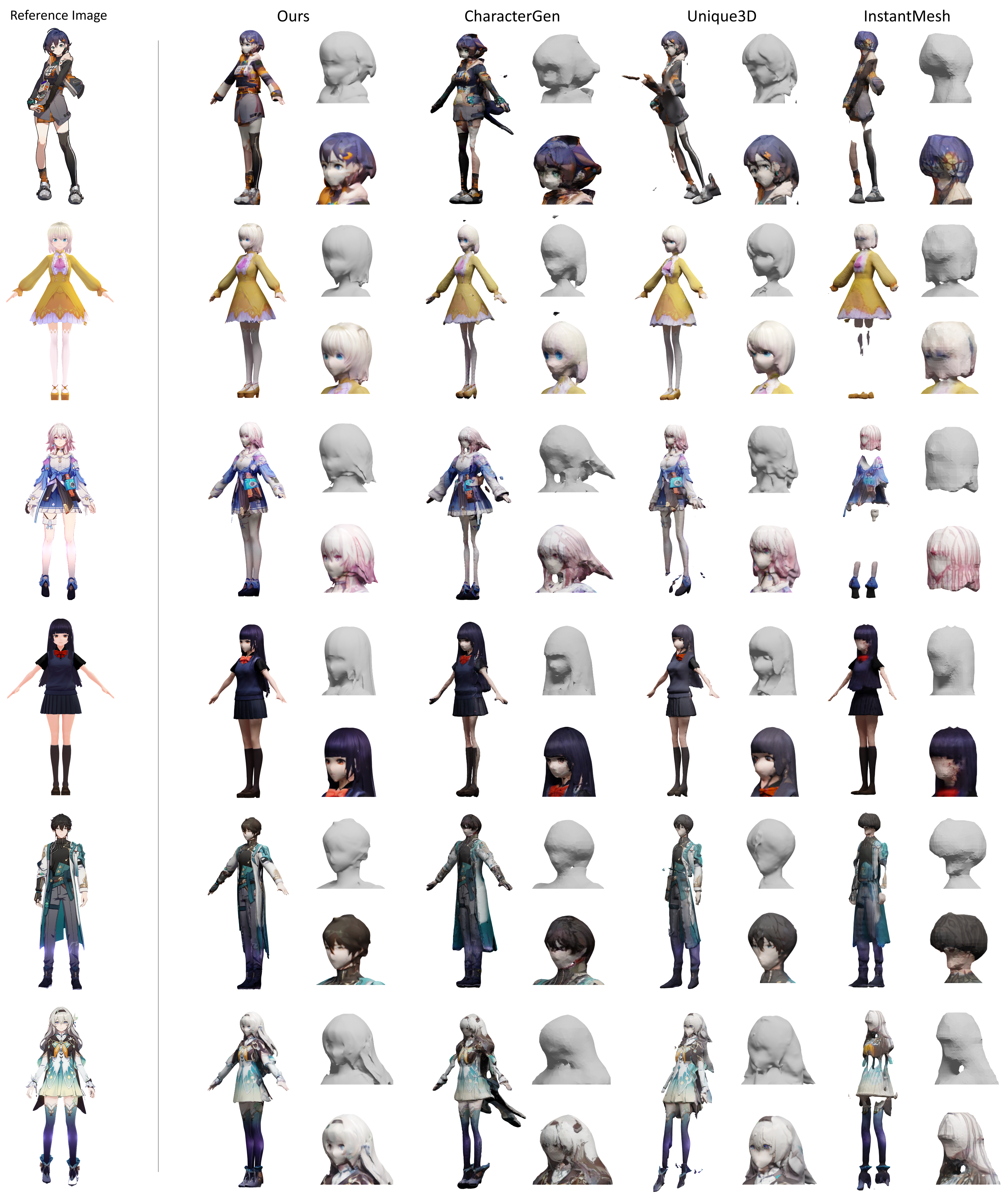}
\caption{More qualitative comparisons of 3D character generations (\#2).}
\label{fig:morequal2}
\end{figure*}

\begin{figure*}[htbp]
\centering
\includegraphics[width=1.0\linewidth]{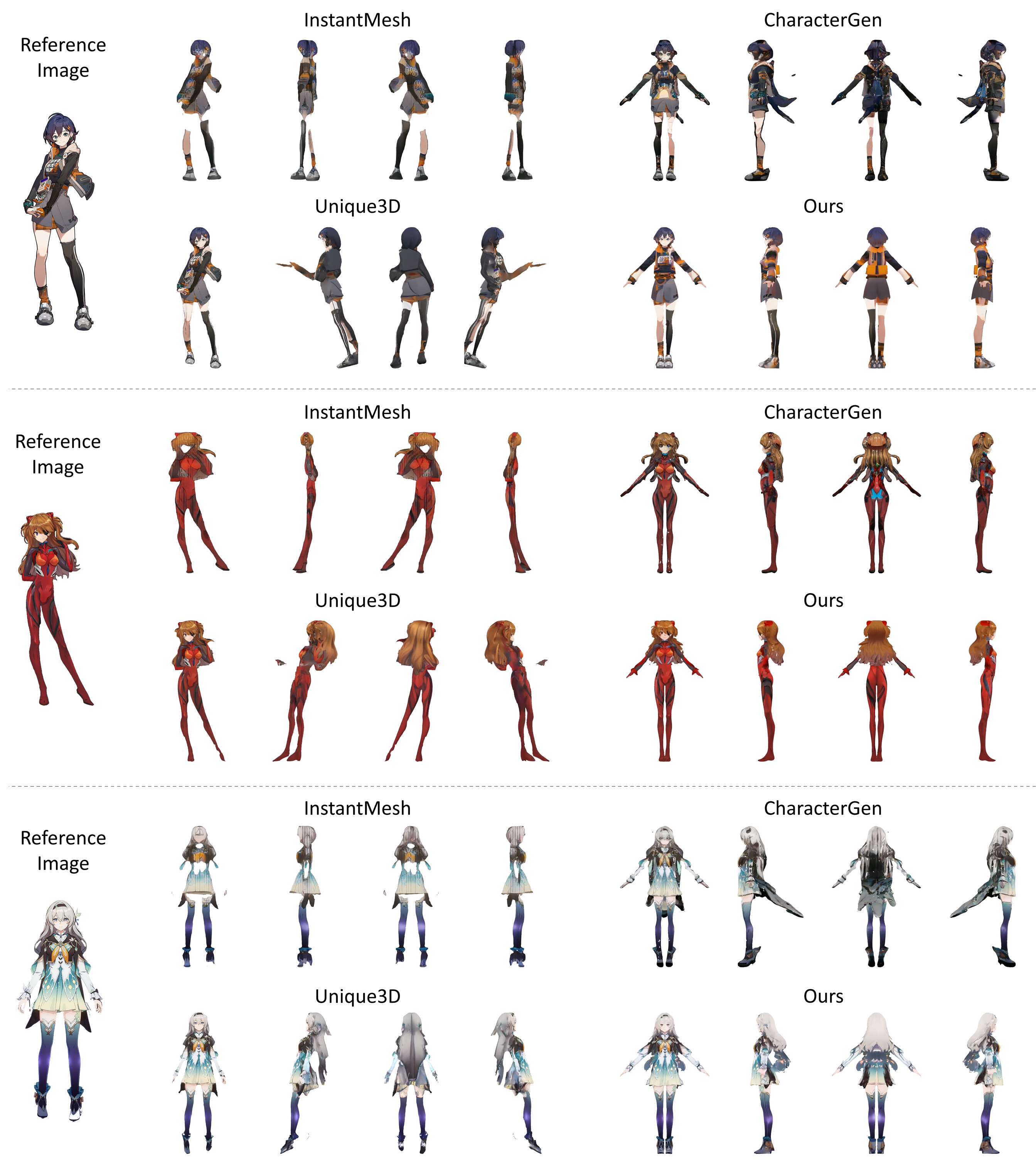}
\caption{More qualitative comparisons of 3D character generations (multi-view renderings).}
\label{fig:multiview}
\end{figure*}

\xhdr{Design choice of Semantic-Equivalent SDF}
Here we discuss why we don't assign a dedicated decoder for each semantic class to predict their SDFs separately. First, this approach would not effectively utilize the prior knowledge from the NeRF training stage. The SDF information predicted by these decoders would differ significantly from what was learned in the previous stage, with each decoder only retaining "its own" portion of the whole original SDF. The semantic information learned during the last stage would also become unusable. In contrast, our method almost completely inherits the prior knowledge from the NeRF training stage and smoothly transitions to the SDF stage without any modifications to the network architecture.

Additionally, this alternative approach would suffer from poor scalability - adding a new semantic class would require adding and training a new triplane feature decoder nearly from scratch and modifying the network structure, while our method does not require relearning geometric information when adding or removing semantics. It would also increase computational and memory costs during both training and inference. Furthermore, without cross-semantic constraints, the surfaces extracted from SDFs of different semantic classes might intersect, which contradicts our requirements. Another issue is that such a representation would not be unified - we could only extract a surface for each individual semantic class, but not for the entire character or multiple selected semantic classes simultaneously. In contrast, our solution can extract equivalent surfaces for any combination of selected semantic classes with only one decoder employed.

\xhdr{Limitations}
We note several limitations that leave room for future work:
(1) Following InstantMesh~\cite{xu2024instantmesh}, our S-LRM produces triplanes with the resolution of $64\times 64$. After switching to SDF training, the FlexiCubes sampling uses a grid size of only 150 height and 100 width. This resolution may constrain and limit further improvement in the results.
(2) While our pipeline enables high-quality generation by the high-resolution diffusion output up to a resolution of 1024, this also slows the overall generation speed, presenting a trade-off. Our S-LRM requires only about 10 seconds for one inference, with the majority of time consumed in the diffusion and refinement, suggesting room for further optimization.
(3) The restricted style and category diversity in the training data affects its ability to handle inputs that deviate significantly from the human-centric categories (e.g., animals or general 3D objects).
Despite such challenges, our framework is extensible, allowing further improvements through tools like SAM for semantic labeling or SMPL label transfer for human datasets.

\section{More Visualizations}

We demonstrate more visualizations of semantic-decomposed 3D results (Fig.~\ref{fig:moredecomp}), outputs of canonicalization (Fig.~\ref{fig:diff1}) and multi-view (Fig.~\ref{fig:diff2}) diffusion model, comparisons with other methods (Fig.~\ref{fig:morequal}, \ref{fig:morequal2}) and multi-view renderings (Fig.~\ref{fig:multiview}).